\documentclass[11pt]{article}

\usepackage[preprint]{acl}

\usepackage{makecell}

\usepackage{algorithm}

\usepackage{algpseudocode}

\usepackage{booktabs}

\usepackage[utf8]{inputenc}
\usepackage[T1]{fontenc}
\usepackage{lmodern}

\usepackage{listings}
\usepackage{tabularx}

\usepackage{times}
\usepackage{latexsym}

\usepackage[T1]{fontenc}

\usepackage[utf8]{inputenc}

\usepackage{microtype}

\usepackage{inconsolata}

\usepackage{graphicx}
\usepackage{url}
\usepackage{multirow}

\usepackage[table,xcdraw]{xcolor}
\usepackage[normalem]{ulem}
\useunder{\uline}{\ul}{}

\usepackage{amsmath}
\usepackage{xcolor}

\usepackage{listings}
\lstdefinelanguage{json}{
  basicstyle=\ttfamily\small,
  showstringspaces=false,
  breaklines=true,
  columns=fullflexible,
  frame=single,
  literate=
   *{0}{{{\color{black}0}}}{1}
    {1}{{{\color{black}1}}}{1}
    {2}{{{\color{black}2}}}{1}
    {3}{{{\color{black}3}}}{1}
    {4}{{{\color{black}4}}}{1}
    {5}{{{\color{black}5}}}{1}
    {6}{{{\color{black}6}}}{1}
    {7}{{{\color{black}7}}}{1}
    {8}{{{\color{black}8}}}{1}
    {9}{{{\color{black}9}}}{1}
}
\usepackage{listingsutf8}

\usepackage{cuted} 

\usepackage{fancyhdr}

\title{CloneMem: Benchmarking Long-Term Memory for AI Clones}

\author{
Sen Hu\textsuperscript{1,$\dagger$,*},
Zhiyu Zhang\textsuperscript{2,$\dagger$},
Yuxiang Wei\textsuperscript{3,$\dagger$},\\
\textbf{Xueran Han}\textsuperscript{4},
\textbf{Zhenheng Tang}\textsuperscript{5},
\textbf{Huacan Wang}\textsuperscript{6},
\textbf{Ronghao Chen}\textsuperscript{1}\\[0.5ex]
\textsuperscript{1}Peking University \quad
\textsuperscript{2}UC Davis \quad
\textsuperscript{3}Georgia Tech \quad
\textsuperscript{4}MBZUAI \quad
\textsuperscript{5}HKUST \quad
\textsuperscript{6}UCAS \\[1.5ex]
\textsuperscript{$\dagger$}Equal contribution \quad
\textsuperscript{*}Corresponding author: \texttt{husen@pku.edu.cn}
}

\begin{document}
\maketitle

\begin{abstract}
AI Clones aim to simulate an individual’s thoughts and behaviors to enable long-term, personalized interaction, placing stringent demands on memory systems to model experiences, emotions, and opinions over time. Existing memory benchmarks primarily rely on user–agent conversational histories, which are temporally fragmented and insufficient for capturing continuous life trajectories. We introduce \textsc{CloneMem}, a benchmark for evaluating long-term memory in AI Clone scenarios grounded in non-conversational digital traces, including diaries, social media posts, and emails, spanning one to three years. \textsc{CloneMem} adopts a hierarchical data construction framework to ensure longitudinal coherence and defines tasks that assess an agent’s ability to track evolving personal states. Experiments show that current memory mechanisms struggle in this setting, highlighting open challenges for life-grounded personalized AI.
Code and dataset are available at \url{https://github.com/AvatarMemory/CloneMemBench}

\end{abstract}
\section{Introduction}
\raggedbottom

In recent years, the rapid development of large language models (LLMs) has given rise to a wide range of personalized applications, such as role-playing agents and AI companions that mimic specific personalities ~\cite{twotalesofpersona}. These applications reveal an underlying trend: users are increasingly seeking to establish deep, personalized connections with AI systems~\citep{personalization_survey2024}. In light of this, researchers have begun to explore a more advanced paradigm— AI Clones, an AI system that can simulate an individual’s thoughts and behaviors~\citep{Text_based_ai_clone, secondme}. AI Clones have substantial practical value, with applications ranging from emotional companionship to large-scale personalized interaction. To support such applications, AI Clone's memory must be carefully designed to capture individual experiences, track changes in emotional states, and model how opinions are formed and evolved over time.

\begin{figure}[t]
  \centering
  \includegraphics[width=\columnwidth]{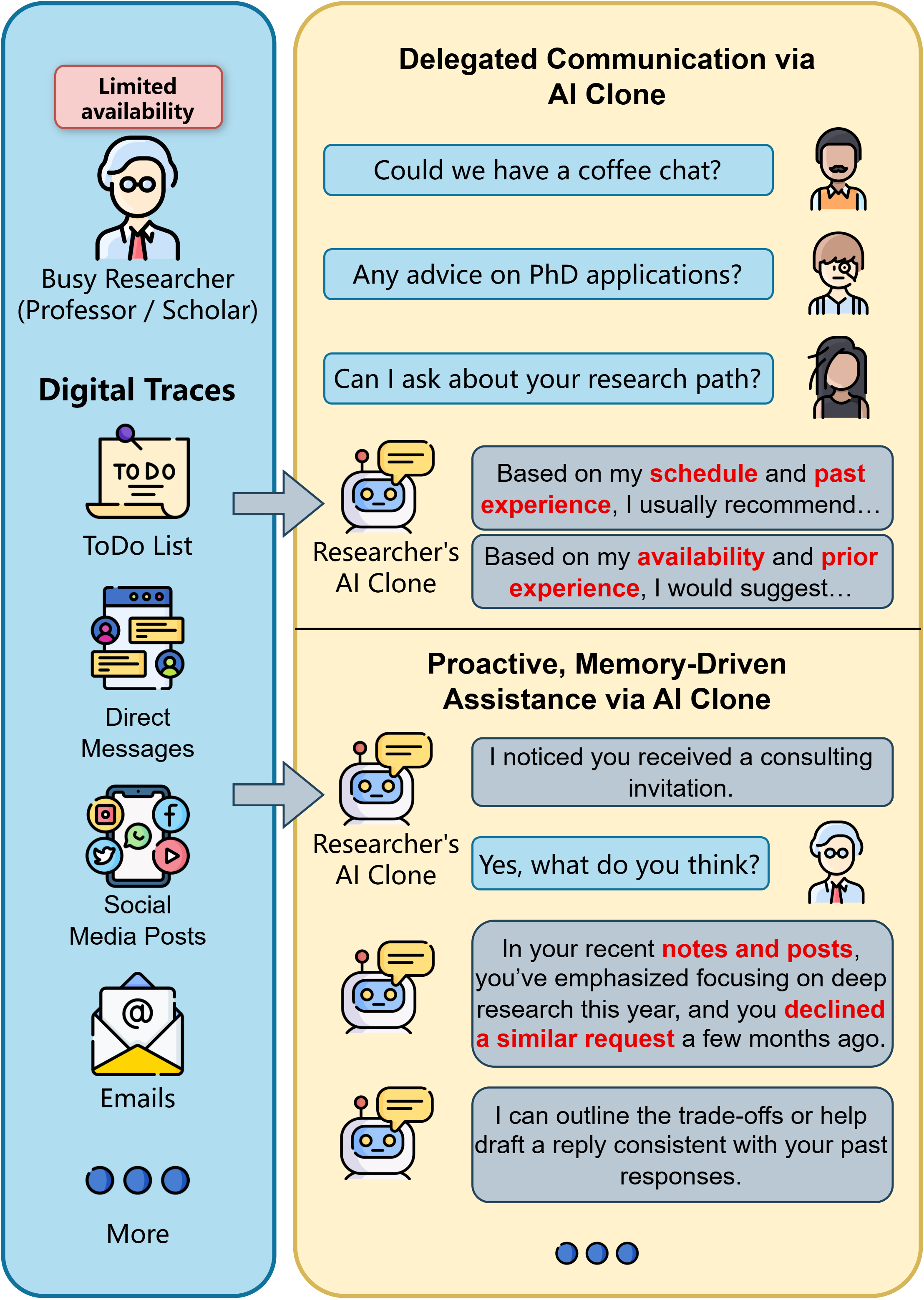}
  \caption{Illustrative application scenarios of an AI Clone grounded in long-term digital traces.}
  \label{fig:ai_clone_scenarios}
\end{figure}

Existing benchmarks for evaluating agent-based user memory are primarily based on interaction paradigms centered on direct user-agent conversational histories \citep{PersonaMem_jiang2025, LoCoMo2024}. These benchmarks assess an agent's ability to maintain information consistency and contextual continuity over extended conversations~\citep{survey_on_memory_mechanism}. As summarized in Table~\ref{tab:benchmark_comparison}, the information used in such benchmarks is derived exclusively from user-agent conversational histories. However, such data are inherently fragmented and episodic in their temporal structure, capturing only sparse snapshots of an individual's experiences. This discrete structure makes it difficult to track gradual changes in psychological states or record everyday life beyond conversational settings. 
Moreover, in real-world settings, injecting memories into an AI Clone through proactive conversational interactions incurs a high time cost.

\begin{table*}[t]
  \centering
  \footnotesize
  \setlength{\tabcolsep}{5pt}
  \begin{tabular}{@{}lclccc@{}}
    \toprule
    \textbf{Benchmark} &
    \textbf{Source} &
    \textbf{Scenario} &
    \textbf{Time Span} &
    \makecell{\textbf{Temporal}\\\textbf{Structure}} &
    \textbf{Consistency} \\
    \midrule
    LoCoMo \citep{LoCoMo2024} &
    CH &
    Chat assistant &
    Several months &
    Discrete sessions &
    Persona \& timeline \\
    LongMemEval \citep{wu2025longmemeval} &
    CH &
    Chat assistant &
    $\sim$2.5 years &
    \makecell{Discrete\\multi-session} &
    Factual \\
    PersonaMem \citep{PersonaMem_jiang2025} &
    CH &
    Chat assistant &
    Several years &
    \makecell{Preference\\snapshots} &
    Preference \\
    \textbf{\textsc{CloneMem} (Ours)} &
    \textbf{DT} &
    \textbf{AI clone} &
    \makecell[l]{\textbf{1$\sim$3 year} \\ \textbf{trajectory}} &
    \makecell[l]{\textbf{Continuous}\\\textbf{life arcs}} &
    \makecell[l]{\textbf{Exp. + Emo. + Opi.}\\\textbf{(longitudinal)}} \\
    \bottomrule
  \end{tabular}
  \caption{\label{tab:benchmark_comparison}
    Comparison of long-term memory benchmarks for personalized AI systems.
    CH = \emph{conversation history}; DT = \emph{non-conversational digital traces};
    Exp. = Experience; Emo. = Emotion; Opi. = Opinion.
  }
\end{table*}

In contrast, the memory of an AI Clone need not be restricted to interactions with the user, but can instead be grounded in digital traces that users naturally generate in their everyday lives, comprising diaries, social media posts, direct messages, emails, etc. Such traces form a naturally occurring, longitudinal record of everyday activities ~\cite{bigfive_digital_footprints}, enabling an AI Clone not only to support more personalized and proactive interactions with its user, but also to serve as a personalized digital proxy that can interact with others while maintaining a coherent and up-to-date user memory. Figure~\ref{fig:ai_clone_scenarios} illustrates such application scenarios.

However, existing benchmarks are not designed to evaluate whether such non-conversational digital traces can be effectively used to track changes in an individual’s experiences, emotions and opinions over time.

We introduce \textsc{CloneMem}, a comprehensive benchmark for evaluating long-term memory capabilities of AI Clone. \textsc{CloneMem} tests whether an AI Clone can integrate non-conversational digital traces drawn from everyday life and use them to consistently track an individual's experiences, emotional changes and evolving opinions over time.

Our main contributions are summarized as follows.
\begin{itemize}
\item We extend memory evaluation beyond traditional conversation-based settings to long-term individual modeling grounded in real-world data. Unlike prior benchmarks that rely only on user--agent conversational histories, our benchmark incorporates non-conversational digital traces as its primary data source.

\item Moving beyond conversation-based settings to non-conversational digital traces introduces new challenges for data curation, as it requires modeling experiences, emotional changes, and evolving opinions over time. We tackle this via a top-down approach: we first generate macro-level life arcs, then derive fine-grained daily traces, ensuring longitudinal coherence in experiences, emotions, and opinions.

\item We design several evaluation tasks to test whether AI Clones can use non-conversational digital traces to track changes in an individual's experiences, emotions, and opinions over time.

\item Our experiments with \textsc{CloneMem} indicate that existing memory systems face limitations in AI Clone scenarios. We analyze possible reasons for these limitations and discuss potential directions for future work.
\end{itemize}

\section{Related Work}

\paragraph{Agent Memory System.}
To overcome the limitations imposed by finite context windows, researchers have proposed various memory systems to enhance long-term memory capabilities in LLM-based agents. Early representative work such as MemGPT ~\citep{MemGPT} treats LLMs as operating systems, using dynamic scheduling between context window and external storage to cope with limited context capacity. This idea later informed a number of memory system designs.

Recent work examines memory design from different angles. HippoRAG \citep{HippoRAG} uses knowledge graphs to support information integration; ReadAgent \citep{ReadAgent} relies on gist-based memory to handle long documents with lower computational cost; Zep \citep{Zep} and Mem0 \citep{Mem0} represent memory as graph structures that can evolve over time; and A-MEM \cite{AMEM} focuses on the dynamic and self-organizing properties of memory.

\paragraph{Personalization and Role Play.}
Recent work has explored role adoption and personality consistency in LLMs.
Character-LLM~\citep{characterllm} and PersonaLLM~\citep{personallm} focus on maintaining predefined roles or personality traits, while AI Persona~\citep{aipersona} examines personalization that develops through ongoing interaction.

Most of the existing work assumes that a person’s persona remains largely unchanged. Our work instead asks whether an AI clone can reflect how a person changes over time as new life events occur.

\paragraph{Long-Term Memory Benchmarks.}
A range of benchmarks have been proposed to evaluate long-term memory and long-context understanding in LLMs.
LoCoMo~\citep{LoCoMo2024}, LongMemEval~\citep{wu2025longmemeval} and MemBench~\citep{membench-2025} focus on multi-session memory, temporal reasoning, and information updating.
Separately, long-context benchmarks such as LongBench v2~\citep{longbenchv2}, BABILong~\citep{BABILong}, and RULER~\citep{ruler2024} evaluate model performance under extremely long inputs.

Despite these advances, existing benchmarks typically rely on single-source data, such as conversational histories or individual documents, making it difficult to capture how an individual’s experiences, emotions, and opinions evolve over time.
\textsc{CloneMem} addresses this gap based on evaluation in non-conversational digital life traces.

\section{Data Construction for \textsc{CloneMem}}

\begin{figure*}[t]
  \centering
  \includegraphics[width=\textwidth]{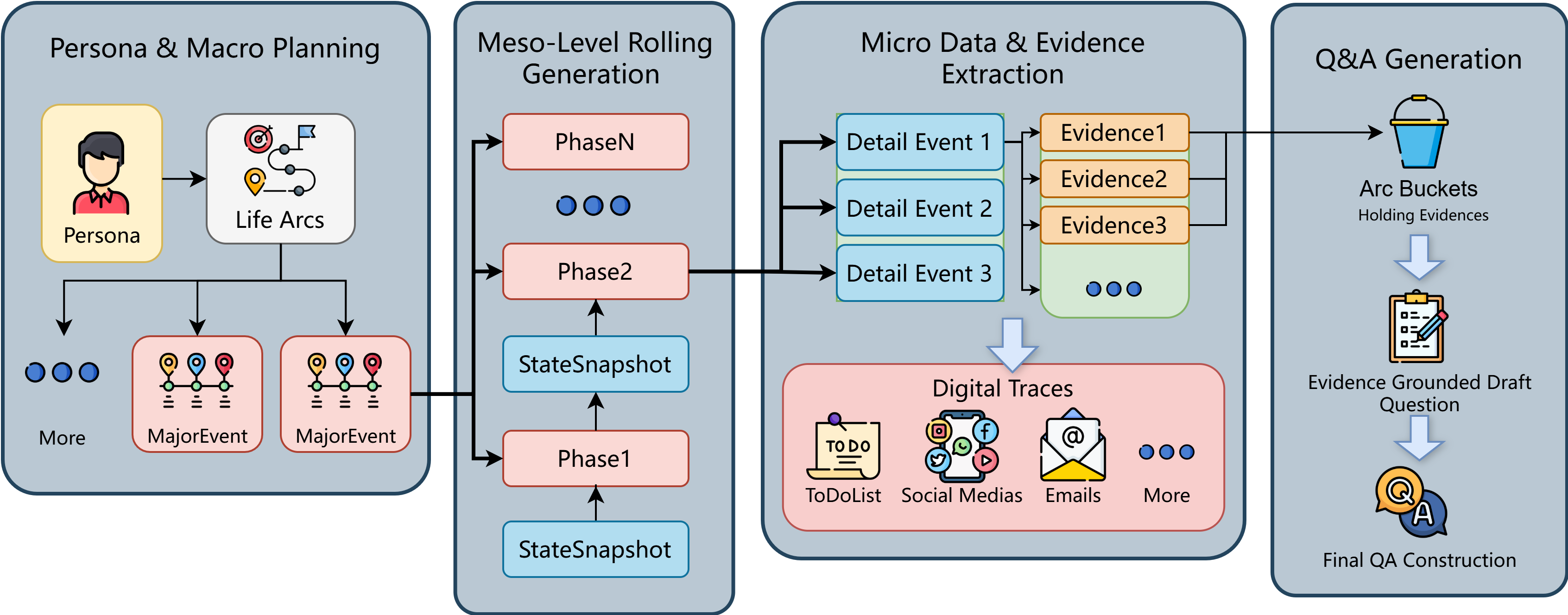}
  \caption{
  \textbf{Overview of the Avatar Memory Benchmark data construction and evaluation pipeline.}
  The pipeline proceeds from persona, macro-level life arcs and major events, through meso-level rolling generation of phases with state snapshots, to micro-level detailed event simulation, evidence extraction and final generation of non-conversational digital traces. Extracted evidence is then organized into arc-specific buckets to support evidence-grounded question--answer generation.
  A detailed version of the pipeline is provided in Appendix~\ref{sec:appendix_A}.
  }
  \label{fig:avatar_memory_pipeline}
\end{figure*}

Moving beyond conversational histories to non-conversational digital traces alters what needs to be modeled. The resulting data form a temporally extended record of an individual’s life rather than a set of isolated utterances. As a result, the modeling focus shifts from detecting whether emotions or opinions change to understanding how such changes arise from prior experiences, which requires experiences, emotions, and opinions to remain coherent over a long time span. To address this, we adopt a hierarchical generation framework \citep{fan-etal-2018-hierarchical_gen}, where high-level life arcs give rise to a structured sequence of major events, phases, and detailed events that define a long-term personal narrative.
Digital traces are generated as non-conversational artifacts grounded in this structure (Figure~\ref{fig:avatar_memory_pipeline}).

\subsection{Hierarchical Event and Trace Generation}
We organize data generation as a three-level hierarchical process, moving from long-term life planning to fine-grained daily digital traces.

\paragraph{Persona and Macro-Level Life Arcs.}
We first construct a psychologically coherent persona based on the five Big Five personality traits ~\citep{big_five_origin}.  
Given this persona, we specify a small set of macro-level life arcs that describe long-term trajectories in experiences, emotions, and opinions.  
Each life arc is anchored by a coarse sequence of major life events, which provide temporal milestones without fixing detailed outcomes.  
As a result, life arcs and major events define a global structure that constrains downstream generation.

\paragraph{Meso-Level Phase Generation.}
At the meso-level, major life events are refined into sequences of phases that mediate between long-term life arcs and daily digital traces. To maintain longitudinal coherence, we associate each phase with a persistent internal state that captures the evolving emotional and attentional context of the persona. This state provides continuity across phases, ensuring that emotional and opinion changes reflect accumulated experiences rather than appearing as isolated signals. By conditioning phase generation on the current internal state, earlier phases are able to influence how subsequent phases develop. (Details of the phase generation process are provided in Appendix~\ref{sec:stage2})

\paragraph{Micro-Level Digital Trace Generation.}
At the micro level, we generate fine-grained daily digital traces that serve as direct inputs to an AI Clone. These traces consist of non-conversational artifacts, including diaries, social media posts, and direct messages. For each phase, we introduce a set of detailed events that represent the concrete activities and experiences of the persona. Both digital traces and explicit evidence are generated with reference to these detailed events, with the evidence encoding the persona’s experiences, emotional states, and expressed or implied opinions. This evidence serves as a grounding constraint during trace generation, ensuring consistency between the resulting traces and the persona’s underlying experiences and internal states (See Appendix~\ref{sec:stage3}).

\subsection{Evaluation Question Construction.}
Based on the generated digital traces and their associated evidence, we construct question–answer (QA) instances to evaluate whether an AI Clone can track how experiences, emotions, and opinions evolve over time.  

Each QA instance is grounded in evidence tied to a specific life arc and reflects a localized temporal span, allowing questions to capture both recent developments and relevant prior context.  

QA instances are generated only when the available evidence is sufficient to support a given question category.  
This evidence-gated construction ensures that each question is well-supported, temporally grounded, and evaluates progression within a life arc rather than isolated facts (See Appendix~\ref{sec:stage4})..

\subsection{Data Quality Control}
We apply a post-processing pipeline that combines automated filtering with limited human review. Automated checks first remove QA instances with formatting errors or missing fields. We then perform an evidence sufficiency check, discarding questions whose referenced evidence does not adequately support the corresponding answers. This scripted filtering step removes approximately 10\% of the generated questions.

After automated filtering, a small subset of the remaining QA instances is reviewed by human annotators to verify plausibility and consistency with the underlying digital traces.
\section{\textsc{CloneMem} Evaluation Benchmark}

Based on the generated digital traces and ground-truth evidence, we introduce the \textsc{CloneMem} benchmark, which features a series of evaluation tasks designed to assess an AI Clone's ability to comprehend and reason about an individual's long-term, evolving life story. Unlike benchmarks that primarily evaluate memory through fact-oriented question answering over conversational history~\citep{LoCoMo2024, wu2025longmemeval}, our tasks are specifically designed to test the ability of AI Clone to track the trajectory of and reasons behind changes in an individual's experiences, emotions, and opinions over time.

In \textsc{CloneMem}, each question is associated with a set of evidence units that specify the semantic unit required to answer correctly. Digital traces are generated from these evidence units and constitute the retrieval space to the AI Clone. Under this formulation, evidence units and digital traces form a many-to-many relationship: a single evidence unit may appear across multiple traces, while a single trace may encode information from multiple evidence units. Therefore, Evaluation in \textsc{CloneMem} is defined over retrieval from digital traces and reasoning with respect to evidence units.

\begin{figure*}[t]
  \centering
  \includegraphics[width=\textwidth]{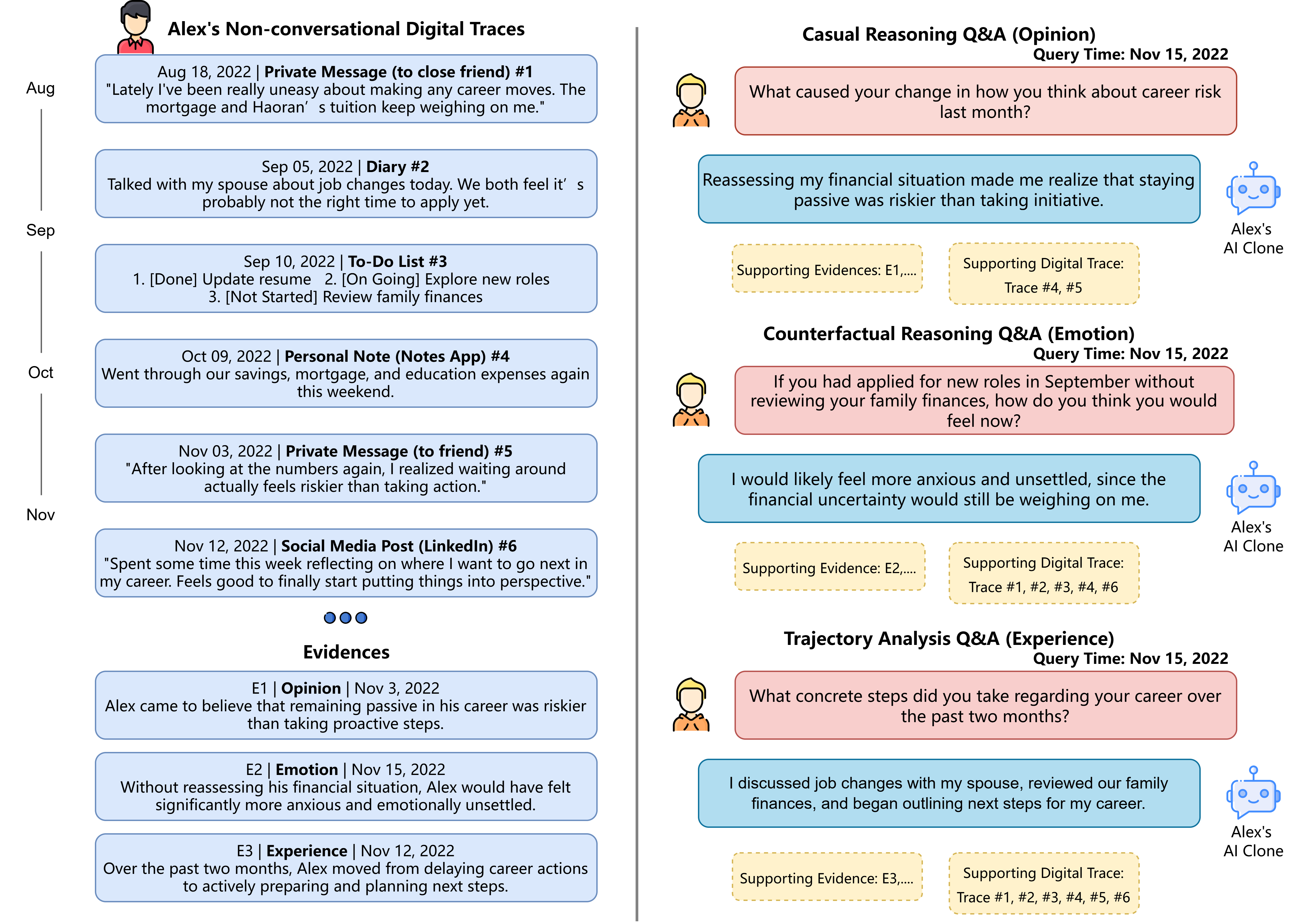}

  \caption{\textbf{Illustrative and representative examples of \textsc{CloneMem} tasks.}
  The left panel shows non-conversational digital traces and their associated ground-truth evidence generated during data construction; the right panel shows example questions and answers for three task types.}
  \label{fig:clonemem_examples}
\end{figure*}

\begin{table}[t]
\centering
\resizebox{0.8\linewidth}{!}{
\begin{tabular}{ll}
\toprule
\textbf{Statistic} & \textbf{Value} \\
\midrule
\# Personas        & 10 \\
\# Questions       & 1,183 \\
Languages          & English, Chinese \\
Context Length     & \makecell[l]{3 short ($\sim$100k tokens),\\ 7 long ($>$500k tokens)} \\
Question Types & \makecell[l]{8 task categories (See Section~\ref{sec:tasks})} \\
\bottomrule
\end{tabular}}
\caption{Overview of dataset statistics.}
\label{tab:dataset_stats}
\end{table}

\subsection{Dataset Statistics}
We generate a bilingual data set (English and Chinese) based on 10 distinct personas. The dataset comprises approximately 5,000 question-answer pairs, with a significant portion of the data designed to test long-context understanding. Specifically, 7 of the 10 personas have context lengths exceeding 500k tokens (some reaching up to 1M), while the remaining 3 are around the 100k token level. Table~\ref{tab:dataset_stats} provides an overview of the dataset statistics.
More detailed breakdowns are reported in Appendix~\ref{sec:stats}.

\subsection{Evaluation Tasks}
\label{sec:tasks}

\textsc{CloneMem} is provided in both free-text and multiple-choice question answering formats, with the latter framing evaluation as a classification task.  
Details of multiple-choice option generation are provided in Appendix~\ref{sec:stage4}.

To reflect realistic long-context interactions, all questions are posed from the perspective of a close friend at a specific point in time.  
Questions often include conversational anchor points that require the AI Clone to locate relevant information within long-term, non-conversational digital traces before answering.

Figure~\ref{fig:clonemem_examples} illustrates representative \textsc{CloneMem} tasks. In general, evaluation tasks are organized around three levels of reasoning: (1) \textbf{factual recall} at specific time points, (2) \textbf{temporal reasoning} about changes, trajectories, and patterns, and (3) \textbf{higher-level reasoning} involving causality, counterfactuals, abstraction, and unanswerable cases.

The formal task definitions and the generation procedures for each type of task are provided in Appendix~\ref{sec:appendix_B} and Appendix~\ref{sec:appendix_A}.

\section{Experimental Setup}

We evaluate memory-augmented systems on the \textsc{CloneMem} benchmark using a standardized retrieval-based evaluation framework. To ensure a fair comparison, all evaluated methods operate on the same long-term, non-conversational digital traces and share a unified pipeline for memory construction, retrieval, and response generation.

\paragraph{Implementation Details.}
Our experiments utilize two backbone language models, LLaMA-3.1-8B \cite{llama3} and GPT-4o-mini, and two embedding models, Contriever \citep{contriever} and text-embedding-3-small. Digital traces are indexed as retrievable memory units using embedding-based similarity search.

At inference time, the system retrieves relevant items based on embedding similarity with a retrieval depth of  $k \in \{5, 10, 20\}$. The backbone models then generate answers based on the retrieved context. For evaluation, we assess free-text responses using GPT-4o as an LLM-based judge, following established protocols \citep{surveyllmjudge}.

\subsection{Baselines}
We compare \textsc{CloneMem} against three paradigms of memory representation, including a non-updatable system (the Flat retriever), and two updatable systems (A-Mem and Mem0). To ensure a fair comparison, all systems are evaluated as retrievers over a pre-processed memory index generated from digital traces, with interactive control loops disabled during the query phase. Additional details on these methods are provided in Appendix~\ref{app:baseline}.

\subsection{Evaluation Metrics}
We evaluate system performance in retrieval, memory utility, and generation quality. For media-level retrieval, we report three variations of Recall@$K$: \textbf{Recall-Flat} measures the overall proportion of unique ground-truth media IDs retrieved; \textbf{Recall-All-Any} provides a strict assessment requiring at least one item from \textit{every} necessary evidence set to be found; and \textbf{Recall-Any-Any} offers a more lenient measure requiring at least one item from \textit{any} evidence set. 

For the quality of model responses, we report \textbf{Choice Accuracy} for multiple-choice tasks. We further employ an LLM-as-a-judge to provide two normalized scores (ranging from 0 to 1): the \textbf{Memory Helpfulness Score}, which evaluates the utility of retrieved traces for answering the query, and the \textbf{QA Consistency Score}, which measures both the truthfulness of the response and its ability to utilize specific personal memories rather than providing generic answers. A detailed breakdown of these metrics and their scoring rubrics is provided in Appendix \ref{app:app_metrics}.

\section{Experiment Results}
In this section, we present the retrieval and LLM-as-judge metrics of exsisiting memory systems tested on \textsc{CloneMem}. More results on question type and dimension analysis and bad case analysis can be found in Appendix \ref{app:qtype}, \ref{app:error}.

\begin{table}[htbp]
\renewcommand{\arraystretch}{1.50}  
\resizebox{\linewidth}{!}{
\centering
\begin{tabular}{c|c|ccc|ccc}
\hline
                      &                  & \multicolumn{3}{c|}{LLaMA-3.1-8b} & \multicolumn{3}{c}{GPT-4o-mini} \\ \cline{3-8} 
                      &                  & A-Mem     & Mem0      & Flat   & A-Mem    & Mem0     & Flat   \\ \hline
\multirow{3}{*}{k=10} & recall\_all\_any & 0.1332    & 0.0838    & \textbf{0.1588}    & 0.2099   & 0.1281   & \textbf{0.2180}    \\
                      & recall\_any\_any & 0.1792    & 0.2914    & \textbf{0.4451}    & \textbf{0.5861}   & 0.4344   & 0.5849    \\
                      & recall\_flat     & 0.1658    & 0.1114    & \textbf{0.1972}    & 0.2687   & 0.1754   & \textbf{0.2752}    \\ \hline
\multirow{3}{*}{k=20} & recall\_all\_any & 0.2173    & 0.1112    & \textbf{0.2489}    & 0.3035   & 0.1934   & \textbf{0.3375}    \\
                      & recall\_any\_any & 0.5494    & 0.3913    & \textbf{0.6103}    & 0.7225   & 0.5515   & \textbf{0.7277}    \\
                      & recall\_flat     & 0.2707    & 0.1588    & \textbf{0.3062}    & 0.3851   & 0.2517   & \textbf{0.3986 }   \\ \hline
\end{tabular}%
}
\caption{Retrieval performance across memory architectures. Bold text indicating the top-performing retriever for each model setting.}
\label{tab:recall_metrics}
\end{table}

\subsection{Main Results}

\paragraph{Retrieval Performance.}
The retrieval results (as in Table \ref{tab:recall_metrics}) demonstrate that the Flat retriever consistently outperforms the more complex updatable memory systems, A-Mem and Mem0, across almost all evaluated metrics and both backbone models. This trend suggests that current abstraction and consolidation techniques, while intended to streamline memory, may inadvertently strip away the fine-grained contextual metadata necessary for precise media-level retrieval. While GPT-4o-mini exhibits stronger retrieval capability than LLaMA-3.1-8B, both models struggle significantly with the Recall-All-Any metric, which requires identifying at least one piece of evidence for every required evidence set. These findings highlight a "lossy compression" trade-off in existing agentic memory frameworks: as information is summarized or consolidated, the link to the original digital traces is weakened, making it harder for the AI Clone to ground its reasoning in specific past experiences.

\begin{table*}[htbp]
\resizebox{\textwidth}{!}{
\centering
\begin{tabular}{c|c|cccc|cccc}
\hline
                       &                            & \multicolumn{4}{c|}{LLaMA-3.1-8b}                          & \multicolumn{4}{c}{GPT-4o-mini}                            \\ \cline{3-10} 
                       &                            & \cellcolor[HTML]{E0E0E0}Oracle & A-Mem  & Mem0   & Flat& \cellcolor[HTML]{E0E0E0}Oracle & A-Mem  & Mem0   & Flat\\ \hline
                       & QA consistency (0-1) & \cellcolor[HTML]{E0E0E0}0.7040 & 0.4829 & 0.4393 & \textbf{0.4971}  & \cellcolor[HTML]{E0E0E0}0.8325 & \textbf{0.6965} & 0.6502 & 0.6955  \\
                       & choice accuracy (0-100)    & \cellcolor[HTML]{E0E0E0}86.06  & 77.91  & \textbf{79.68}  & 78.76   & \cellcolor[HTML]{E0E0E0}89.65  & 87.48  & 85.28  & \textbf{87.74}   \\
                       & memory helpful score (0-1) & \cellcolor[HTML]{E0E0E0}0.9333 & 0.4937 & 0.4248 & \textbf{0.5721}  & \cellcolor[HTML]{E0E0E0}0.9351 & 0.6275 & 0.6182 & \textbf{0.7767}  \\
\multirow{-4}{*}{k=5}  & memory recall (0-1)        & \cellcolor[HTML]{E0E0E0}0.9994 & 0.4303 & 0.3060 & \textbf{0.4758}  & \cellcolor[HTML]{E0E0E0}1      & 0.5490 & 0.4860 & 0.6492  \\ \hline
                       & QA consistency (0-1) & \cellcolor[HTML]{E0E0E0}0.6892 & \textbf{0.4966} & 0.4384 & 0.4730  & \cellcolor[HTML]{E0E0E0}0.8368 & \textbf{0.7195} & 0.6709 & 0.7187  \\
                       & choice accuracy (0-100)    & \cellcolor[HTML]{E0E0E0}86.03  & \textbf{78.65}  & 72.43  & 77.81  & \cellcolor[HTML]{E0E0E0}89.25  & 88.06  & 87.64  & \textbf{88.50}   \\
                       & memory helpful score (0-1) & \cellcolor[HTML]{E0E0E0}0.9321 & 0.4971 & 0.4270 & \textbf{0.5748}  & \cellcolor[HTML]{E0E0E0}0.9331 & 0.6357 & 0.6204 & \textbf{0.7781 } \\
\multirow{-4}{*}{k=10} & memory recall (0-1)        & \cellcolor[HTML]{E0E0E0}1      & 0.4384 & 0.3128 & \textbf{0.4807}  & \cellcolor[HTML]{E0E0E0}0.9994 & 0.5557 & 0.5008 & \textbf{0.6530 } \\ \hline
                       & QA consistency (0-1) & \cellcolor[HTML]{E0E0E0}0.6910& 0.4285& \textbf{0.4384} & 0.4214& \cellcolor[HTML]{E0E0E0}0.8302& 0.7241& 0.6901& \textbf{0.7428}\\
                       & choice accuracy (0-100)    & \cellcolor[HTML]{E0E0E0}85.86& 68.53& \textbf{72.43}  & 69.20& \cellcolor[HTML]{E0E0E0}89.08& 87.63& \textbf{88.01}& 87.13\\
                       & memory helpful score (0-1) & \cellcolor[HTML]{E0E0E0}0.9329& 0.5051& 0.4270 & \textbf{0.5765}& \cellcolor[HTML]{E0E0E0}0.9341& 0.6493& 0.6209& \textbf{0.7769}\\
\multirow{-4}{*}{k=20} & memory recall (0-1)        & \cellcolor[HTML]{E0E0E0}0.9989& 0.4463& 0.3128 & \textbf{0.4845}& \cellcolor[HTML]{E0E0E0}0.9988& 0.5736& 0.5024& \textbf{0.6508}\\ 
\hline
\end{tabular}%
}
\caption{LLM-based evaluation of QA performance and memory utility. Results compare memory systems against an Oracle baseline (ground-truth context). Bold values denote the top-performing retriever excluding the Oracle.}
\label{tab:llm_metrics}
\end{table*}

\paragraph{QA Performance.}
In Table \ref{tab:llm_metrics}, we further show the QA performance and evaluate the memory accuracy. We include the oracle setting, which directly uses the ground-truth original context and extracted statement to answer each question. Across both backbones, the Oracle setting forms a clear upper bound, indicating that most remaining errors come from imperfect memory construction/retrieval rather than the task itself.

Among non-Oracle memory systems, Flat retrieval is the most reliable for semantic memory utility: it achieves the best (or near-best) memory helpfulness and memory recall for both backbones at all k, and improves slightly as k increases. This suggests that aggressive memory consolidation/organization (Mem0, A-Mem) doesn ot reliably help on \textsc{CloneMem}. However, higher recall/helpfulness does not always translate into better QA: for LLaMA-3.1-8B, increasing k yields only marginal gains in recall/helpfulness but can reduce QA consistency and choice accuracy (notably at k=20), consistent with noise/irrelevant retrieval overwhelming a weaker reasoner. GPT-4o-mini is more robust: QA consistency remains high and tends to improve with larger k, and A-Mem is competitive (often best at k=5/10 for consistency), implying that hierarchical abstraction can help a strong model focus on higher-level trajectory signals when retrieval depth is limited.

\begin{table}[htbp]
\centering
\resizebox{\linewidth}{!}{
\begin{tabular}{c|cccc}
\hline
\textbf{Metric} & \textbf{LLaMA/} & \textbf{LLaMA/} & \textbf{4o-mini/} & \textbf{4o-mini/} \\
 & \textbf{contriever} & \textbf{text-emb-3-s} & \textbf{contriever} & \textbf{text-emb-3-s} \\ \hline
recall\_all\_any      & 0.1588 & \textbf{0.3059} & 0.1725 & \underline{0.2180} \\
recall\_any\_any      & 0.4451 &\textbf{0.7029} & 0.4754 & \underline{0.5849} \\
recall\_flat          & 0.1972 & \textbf{0.3755} & 0.2183 & \underline{0.2752} \\ \hline
QA consistency        & 0.4730 & 0.5281 & \underline{0.6946} & \textbf{0.7187} \\
choice accuracy       & 77.81  & 72.91  & \underline{86.06}  & \textbf{88.50}  \\
memory helpful score  & 0.5748 & \underline{0.7402} & 0.6549 & \textbf{0.7781} \\
memory recall         & 0.4807 & \underline{0.6132} & 0.5542 & \textbf{0.6530} \\ \hline
\end{tabular}%
}
\caption{Ablation study of embedding models and backbone LLMs with the Flat retriever. Bold denotes the best performance; underline denotes the second best.}\label{tab:ablate}
\end{table}

\subsection{Ablation Studies}

\paragraph{Embedding and Extraction Model Choice.}
To investigate the impact of the underlying architectures on \textsc{CloneMem} performance, Table \ref{tab:ablate} examines the impact of embedding (\textit{Contriever} vs. \textit{text-embedding-3-small}) and LLM backbone (\textit{LLaMA-3.1-8B} vs. \textit{GPT-4o-mini}) choices. While \textit{text-embedding-3-small} nearly doubles retrieval recall, this does not guarantee superior downstream performance. Notably, \textit{GPT-4o-mini} paired with a weaker retriever outperforms \textit{LLaMA-3.1-8B} with a strong one on consistency and accuracy. This highlights a clear division of labor: embeddings determine the retrieval floor, but the backbone’s reasoning capacity sets the ceiling for final response quality.

\paragraph{Impact of Retrieval Unit Composition.}
Table \ref{tab:ablate_retr} ablates the composition of retrieval units by comparing the indexing of combined data against extracted memories only (\textit{w/o org}) and raw context only (\textit{w/o mem}). A distinct divergence emerges between retrieval metrics and downstream task performance: while systems utilizing extracted memories (\textit{w/o org}) achieve the highest semantic recall and helpfulness scores due to the informational density of summaries, this does not translate to superior reasoning accuracy. Conversely, relying solely on raw original context (\textit{w/o mem}) yields the highest Choice Accuracy, notably outperforming the combined baseline. This highlights a critical "validity-fidelity" trade-off; extracted memories act as effective semantic indices for locating general topics (validity) but suffer from lossy compression, whereas raw digital traces retain the granular fidelity essential for the precise trajectory tracking required in \textsc{CloneMem}.

\begin{table}[]
\renewcommand{\arraystretch}{1.20}  
\centering
\resizebox{\linewidth}{!}{
\begin{tabular}{lcccccc}
\toprule
                     & \multicolumn{3}{c}{k=10}                & \multicolumn{3}{c}{k=20}                 \\ 
\cmidrule(lr){2-4} \cmidrule(lr){5-7}
                     & flat   & flat (w/o org) & flat (w/o mem) & flat   & flat (w/o org) & flat (w/o mem) \\ 
\midrule
recall\_all\_any     & 0.1588 & \textbf{0.1638}         & 0.1400         & \textbf{0.2489} & 0.2476         & 0.1991         \\
recall\_any\_any     & \textbf{0.4451} & 0.4398         & 0.3860         & \textbf{0.6103} & 0.5910         & 0.5416         \\
recall\_flat         & \textbf{0.1972} & 0.1942         & 0.1629         & \textbf{0.3062} & 0.2949         & 0.2565         \\ 
\midrule
QA consistency       & 0.4730 & \textbf{0.4811}         & 0.4715         & 0.4214 & 0.3984         & \textbf{0.4299}         \\
choice accuracy      & 77.81  & 76.05          & \textbf{79.67}          & 69.20  & 69.50          & \textbf{85.98}          \\
memory helpful score & 0.5748 & \textbf{0.6443}         & -              & 0.5765 & \textbf{0.6389}         & -              \\
memory recall        & 0.4807 & \textbf{0.5186}         & -              & 0.4845 & \textbf{0.5252}         & -              \\ 
\bottomrule
\end{tabular}%
}
\caption{Compare indexing combined data, extracted memories only (\textit{w/o org}), and raw context only (\textit{w/o mem}). While extracted memories improve semantic retrieval metrics, raw context is critical for maximizing downstream accuracy.}
\label{tab:ablate_retr}
\end{table}

\section{Discussion: When Existing Memory Systems Fail for AI Clones}
\label{sec:discussion}

\paragraph{Abstraction helps \emph{search}, but hurts \emph{cloning}.}
Across backbones and metrics, the simplest \textbf{Flat retriever} is consistently the most reliable non-oracle baseline for both media-level recall and semantic utility, while consolidation-based memories (Mem0, A-Mem) often underperform despite being designed to ``organize'' history. This pattern supports a core mismatch: AI-clone queries in \textsc{CloneMem} are not satisfied by topic-level recall alone, but require \emph{high-fidelity grounding} in the original traces (timestamps, phrasing, repetition, and cross-event linkage). Summarization and fact extraction act as a lossy compression that weakens the alignment between retrieved memory and the underlying evidence units, improving semantic indexing in some cases but degrading precise trace-level retrieval and trajectory tracking. The ablations reinforce this validity--fidelity trade-off: extracted memories can raise semantics-aware recall/helpfulness, yet raw traces more often preserve the details needed for correct decisions.

\paragraph{When evidence is underspecified, models fall back to narrative priors.}
Even when retrieval surfaces relevant context, the generation step frequently substitutes a coherent story for the true mechanism of change. The emotion case study (\S\ref{sec:case_study_emotion}) illustrates a recurring failure mode: models prefer high-probability \textbf{narrative templates} (e.g., a child-triggered epiphany) over the ground-truth internal pivot, producing fluent but fabricated triggers. This is not merely a retrieval error; it reflects that common memory schemas emphasize \emph{events and interactions} (who did what) while failing to preserve the \emph{belief update} that explains why later behavior differs. For AI Clones, this is a reliability risk: answers can be emotionally plausible yet causally wrong, because the system optimizes for narrative coherence rather than evidential faithfulness.

\paragraph{Event logs cannot represent ``no decision yet.''}
\textsc{CloneMem} also exposes a second conceptual gap: \textbf{activity $\neq$ state}. In the experience case study (\S\ref{sec:case_study_experience}), dense traces of job-search behavior are misread as commitment, leading to ``safe hallucinations'' that invent preferences or actions in questions labeled unanswerable. This suggests that event-centric memory (e.g., ``searched X'') is insufficient for clones, which must model \textbf{state persistence} (e.g., prolonged indecision) and maintain the discipline to answer ``not specified'' even under heavy, noisy retrieval. More broadly, our results show a division of labor: better embeddings raise the retrieval floor, but stronger backbones are required to resist narrative completion and to maintain correct abstention under ambiguity.

\paragraph{Implication.}
Memory for AI Clones should be designed less as a compact knowledge base and more as an evidence-preserving substrate that (i) retains trace-level fidelity, (ii) explicitly represents internal-state transitions (belief/goal shifts) alongside events, and (iii) supports abstention via persistent-state modeling when the record does not warrant a conclusion.

\section{Conclusion}
We introduce \textsc{CloneMem} to evaluate AI Clones to model the evolution of an individual’s opinions, emotions, and experiences using non-conversational longitudinal digital traces. Our experiments show that simple flat retrieval often outperforms abstractive memory systems, which tend to discard critical temporal cues when compressing evidence, leading to a trade-off between validity and fidelity. Our analysis further reveals that AI Clones often depend on generic \textit{narrative templates} when generating responses and struggle to distinguish exploratory activities from commitments. Our findings highlight the need for memory architectures that can model an individual's longitudinal trajectory, capturing not only what changes over time but also what brings to these changes, while preserving fidelity to the underlying traces rather than compressing them away. We position \textsc{CloneMem} as a benchmark to facilitate future research on reliable memory for AI Clones.

\section*{Limitations}

First, while our hierarchical generation pipeline ensures privacy and longitudinal coherence, the resulting digital traces are synthetic. They may lack the chaotic irregularity, noise, and specific linguistic idiosyncrasies found in real-world user data, potentially simplifying the retrieval challenge compared to organic environments. Second, \textsc{CloneMem} currently represents non-textual artifacts, such as photos and voice notes, through textual descriptions. This abstraction bypasses the challenges of native multimodal processing, which remains essential for a fully holistic AI Clone. Third, our evaluation relies on an LLM-as-a-judge framework. Despite the use of rigorous rubrics and safe baselines, scoring memory utility and consistency via GPT-4o may still introduce subtle biases or fail to capture the full nuance of human judgment in ambiguous scenarios. Finally, with ten distinct personas, the benchmark covers a specific range of personality traits and life trajectories but may not fully represent the vast cultural, linguistic, and behavioral diversity necessary for a universally applicable system.


\bibliography{custom}
\newpage
\appendix
\section{Appendix: Data Construction Pipeline}
\label{sec:appendix_A}
We construct the \textsc{CloneMem} benchmark through a multi-stage process that generates digital traces, structured evidence, and question–answer pairs. This appendix focuses on explaining the data construction process rather
than on implementation details.

\subsection{Overview}
\begin{figure*}[t]
  \centering
  \includegraphics[width=\textwidth]{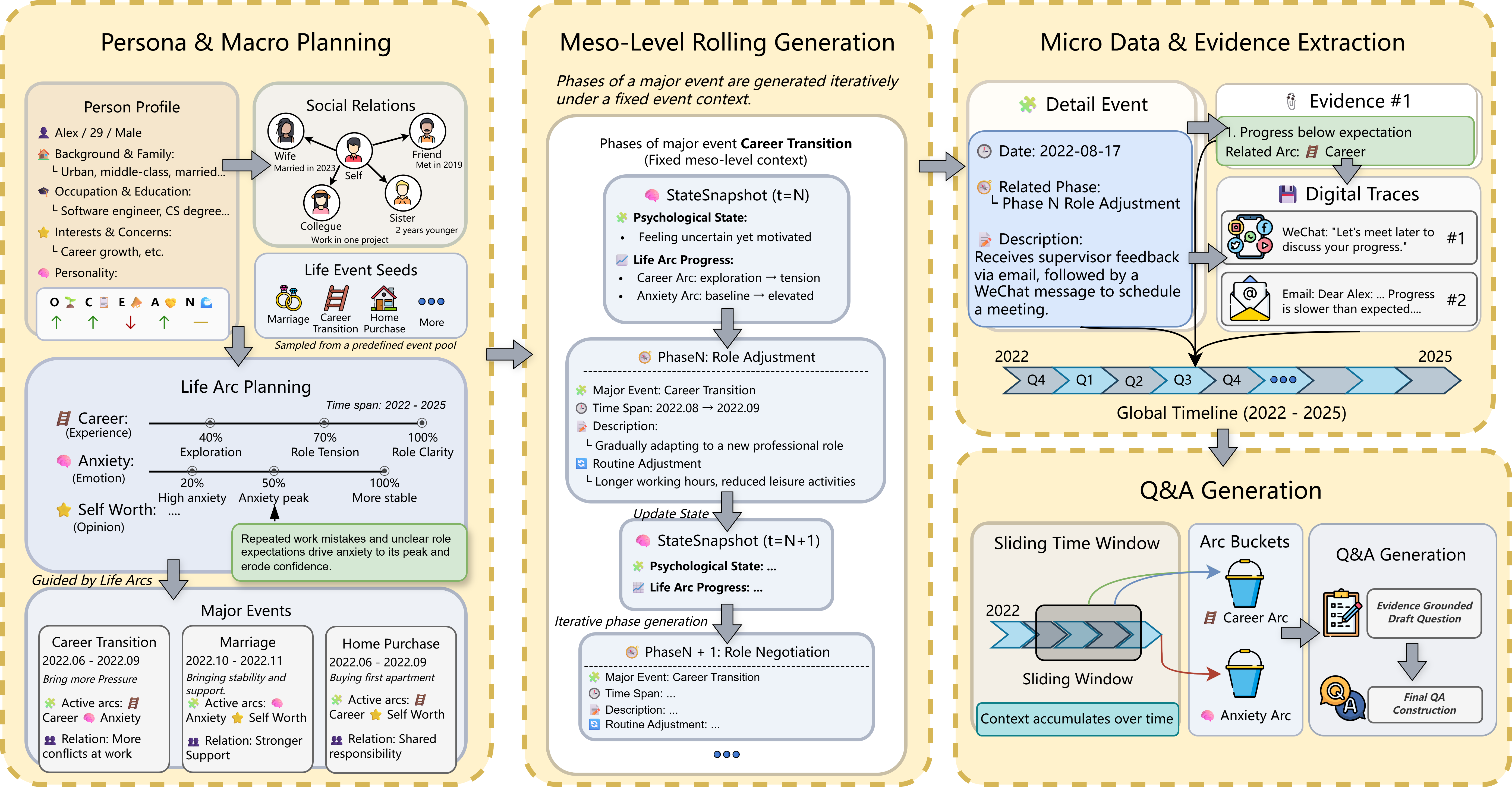}
  \caption{High-level illustration of the \textsc{CloneMem} data construction pipeline. Starting from persona initialization and macro-level life planning, the pipeline expands predefined and model-augmented event seeds into life arcs, major events, phases, and fine-grained detailed events. Each detailed event is generated together with explicit evidence, which jointly ground the generation of non-conversational digital traces. Finally, evidence is aggregated over sliding time windows and life arcs
to construct temporally grounded QA instances.}
  \label{fig:illustrative_pipeline}
\end{figure*}

\textsc{CloneMem} is constructed through a four-stage data construction pipeline, as illustrated in Figure~\ref{fig:illustrative_pipeline}. The pipeline
progressively refines abstract personality traits into temporally grounded digital traces and trajectory-based question--answer instances, while preserving long-term coherence in experiences, emotions, and opinions.

In the first stage (\emph{Persona and Macro-Level Planning}), each persona is initialized from a fixed Big Five personality configuration and enriched
into a structured background profile. Based on this profile, persistent social relations and a set of macro-level life arcs are constructed to define long-term trajectories of opinions, experiences, and emotional
baselines. The major life events are then positioned along the global timeline and aligned with these life arcs, providing a high-level narrative structure
without prescribing fine-grained outcomes.

In the second stage (\emph{Meso-Level Rolling Generation}), each major life event is decomposed into a sequence of phases generated under a rolling
snapshot mechanism. An explicit internal snapshot is maintained and updated across phases and events, allowing accumulated experiences and emotional states to influence subsequent generation. This stage bridges long-term
life arcs and short-term experiences by enforcing temporal continuity at the meso-level.

In the third stage (\emph{Micro-Level Event, Evidence, and Trace Generation}), each phase is further expanded into detailed events. For each detailed event, explicit evidence entries are generated alongside the event itself,
which jointly ground the construction of non- conversational digital traces such as diaries, social media posts, and direct messages. These micro-level
artifacts form the fine-grained record of the evolving life of the persona.

In the fourth stage (\emph{Trajectory-Based Question-Answer Generation}),
evidence is organized along life-arc-specific trajectories using a
sliding-window mechanism over time. Accumulated evidence within each
trajectory segment is then used to construct question--answer instances
that probe how experiences, emotions, and opinions evolve, enabling
evaluation of temporal and trajectory-based reasoning.

Stages I–III of the data construction pipeline are implemented using Claude-4.5-Haiku, covering persona enrichment, life arc generation, and digital trace construction. Stage IV (trajectory-based question–answer generation and multiple-choice construction) is implemented using Gemini-3-Flash (preview).

\subsection{Stage I: Persona and Macro-Level Planning}
\label{sec:stage1}
Each persona in \textsc{CloneMem} is initialized with a fixed configuration of Big Five personality traits~\citep{big_five_origin}. Although these traits provide a stable psychological prior, they are too abstract to directly support event-level generation. We therefore perform a persona enrichment step using an LLM to expand the trait configuration into a structured persona profile. In practice, the enriched profile specifies basic attributes (e.g., name, age, gender),
a natural-language personality description consistent with the Big Five traits, and a biographical context such as family history, education, occupation, and the recurring concerns and interests of the persona. In addition, we include a short narrative description of the persona's life before 2022, which provides the temporal starting point for all subsequent generations.

The enriched persona profile also allows us to identify persistent social relations (e.g., family members, partners, colleagues) that recur across the generated timeline.

All life events, phases, and digital traces generated in \textsc{CloneMem} occur after 2022. The temporal span of post-2022 generation is configurable and varies across personas to support different data scales, ranging from shorter timelines to multi-year personal histories. This design allows \textsc{CloneMem} to evaluate memory systems in moderate and extremely long context settings; the resulting distribution of timeline lengths and data scale is summarized in Table~\ref{tab:dataset_stats}.

To capture long-term personal development, we build a set of macro-level life arcs that define how the persona's experiences, opinions, and emotions evolve over time. Life arc generation is guided by a pool of event seeds that specify the types of event the persona may
encounter. This seed pool is assembled in two stages: first, we maintain a curated set of predefined life event concepts (e.g., graduation, career transition, marriage, breakup) specified in a local configuration; second,
an LLM filters these concepts for persona compatibility and introduces additional persona-specific event seeds to increase diversity.

Given the resulting event seeds, the model generates multiple life arcs spanning three dimensions: opinion, emotion, and experience trajectories. Each life arc is represented by a sequence of anchor states that describe its progression, trigger logic, and observable behavioral cues. Major life events are then instantiated along
the global timeline and aligned with the active life arcs, recording the anchor state that each arc occupies at the time of the event. This alignment provides high-level narrative constraints while leaving room for variability at finer temporal scales.

\subsection{Stage II: Meso-Level Rolling Generation}
\label{sec:stage2}
At the meso level, each major life event is decomposed into a sequence of phases that represent gradual transitions in the persona's experiences and psychological state, rather than isolated incidents. Phase generation is governed by a rolling mechanism with an explicit internal state, which ensures temporal coherence both within a major event and across successive
events.

Before generating the phases of a major event, we initialize a phase snapshot that summarizes the current condition of the persona. This snapshot captures coarse-grained but persistent signals, including energy level, stress level, dominant emotion, current attentional focus, and the status of active life arcs at that point in time. The phases are then generated sequentially. After each phase is produced, the snapshot is updated to
reflect the changes induced by that phase, and the updated snapshot is used as contextual input to generate the next phase.

In particular, the final snapshot of a major life event is not discarded. Instead, it is propagated as the initial snapshot for the subsequent major event. In this way, accumulated experiences and emotional states directly influence how later events unfold. This rolling generation process prevents phases from being treated as independent narrative segments and forces macro-level life arcs to be consistently grounded in meso-level narrative progression.

\begin{algorithm}[t]
\caption{Rolling phase generation with a carried snapshot}
\label{alg:rolling_phase}
\begin{algorithmic}[1]
\Require Ordered major events $\{E_1, \dots, E_N\}$, initial phase snapshot $H_0$
\State $H \gets H_0$
\For{event $E$ in $\{E_1, \dots, E_N\}$}
    \State initialize empty list $\mathcal{P}(E)$
    \For{$k = 1$ to $\textsc{NumPhases}(E)$}
        \State $\textit{phase} \gets \textsc{GeneratePhase}(E, H)$
        \State append $\textit{phase}$ to $\mathcal{P}(E)$
        \State $H \gets \textsc{UpdateSnapshot}(H, \textit{phase})$
    \EndFor
\EndFor
\State \Return $\{\mathcal{P}(E_1), \dots, \mathcal{P}(E_N)\}$
\end{algorithmic}
\end{algorithm}

\subsection{Stage III: Micro-Level Event, Evidence, and Trace Generation}
\label{sec:stage3}
At the micro level, each phase is expanded into a set of detailed events that describe concrete, day-to-day experiences occurring within the phase. These detailed events provide the finest temporal granularity in the
generation pipeline and serve as the immediate basis for evidence and digital trace construction.

For each detailed event, we generate a small set of explicit evidence entries that encode its core factual content, emotional signals, and expressed opinions. Importantly, evidence is generated jointly with the
detailed event rather than extracted post hoc from downstream text. This co-generation design ensures that evidence, events, and traces remain aligned in semantic content and narrative intent, while avoiding information
loss that can arise from later summarization.

Conditioned on the detailed event, its associated evidence, and the historical phase snapshots, we generate non-conversational digital traces such as diary entries, social posts, emails, and private messages. The generation of trace follows a two-step procedure. The model first determines the communicative intent and audience of the trace, as well as the key information to be conveyed. The trace content is then generated under these constraints, grounding surface-level expressions in both the local event context and the evolving psychological state of the persona.

All micro-level artifacts--including detailed events, evidence entries, and digital traces--are stored together with their temporal and semantic relations in a structured graph representation. This representation
preserves fine-grained connections across events, phases, and life arcs, and supports downstream aggregation over time windows and arc-specific contexts for question construction.

\subsection{Stage IV: Trajectory-Based Question-Answer Generation}
\label{sec:stage4}
Based on the generated evidence and digital traces, we construct question-answer (QA) instances that probe an AI Clone's ability to reason about how experiences, emotions, and opinions evolve over time. Rather than
treating evidence as an unordered set, QA generation in \textsc{CloneMem} is grounded in temporally structured trajectories aligned with long-term life arcs.

Each evidence entry is associated with one or more life arcs defined in Stage~I. We organized the evidence into arc-specific buckets, where each bucket collects evidence relevant to a particular opinion, experience, or
emotional trajectory. To model gradual change, we adopt a sliding-window mechanism over the global timeline. As the window moves forward in time, new evidence is incrementally added to the corresponding arc buckets,
allowing information to accumulate and trajectories to emerge.

At each window position, we identify the set of active buckets whose contents have changed due to newly added evidence. The aggregated evidence within an active bucket forms a localized trajectory segment for a given
life arc, reflecting the state of the persona and its recent evolution along that dimension. These trajectory segments serve as the grounding context for QA construction.

QA generation itself follows a two-step procedure. First, for each type of question, the model determines whether a meaningful question can be generated from the available trajectory segment. If so, it plans the question by
selecting relevant evidence, deciding the temporal scope, and drafting a question outline tailored to the type of target reasoning. Second, the model generates the final question and its corresponding answer grounded in the
selected evidence.

To ensure data quality, we apply a post-processing step to remove invalid QA pairs, including cases with formatting errors or misaligned question–answer content. The remaining free-text QA pairs are then converted into multiple-choice variants.

During this conversion, distractor options are generated and iteratively refined through a validation process. One model is used to construct candidate distractors, while a second model is used to assess whether the resulting multiple-choice question can be answered without relying on temporal or trajectory-level reasoning. Questions that are found to be solvable through superficial cues or single-point recall are revised by regenerating their distractors. This process continues until the multiple-choice question requires reasoning over the temporal structure or trajectory-level evidence.

In our implementation, Gemini-3-Flash (preview) is used for distractor generation and GPT-4o-mini for validation during multiple-choice construction; GPT-4o-mini is also used for evaluation, but evaluation signals are not used in the construction of data sets.

\section{Evaluation Tasks}
\label{sec:appendix_B}
\textsc{CloneMem} is released in both free-text and multiple-choice question answering (QA) formats.  
The multiple-choice setting formulates the evaluation as a classification problem, with details of option construction provided in Appendix~\ref{sec:stage4}.

To approximate realistic long-context interactions, all questions are asked from the perspective of a close friend at a specific point in time.  
Questions are grounded in shared experiences and often contain conversational cues, requiring the AI Clone to identify and integrate relevant information from long-term, non-conversational digital traces before producing an answer.

The evaluation tasks in \textsc{CloneMem} span multiple levels of reasoning, ranging from basic factual recall to more abstract forms of inference.  
Specifically, the tasks cover: 
(1) retrieval of factual information at specific time points, 
(2) reasoning over temporal change and continuity, and 
(3) higher-level reasoning that involves causality, hypothetical alternatives, abstraction across experiences, and the ability to recognize missing evidence.
We describe each category in the following.

\paragraph{Single-Point Factual Reasoning}
This task assesses whether an AI Clone can retrieve explicit factual information about an individual's state, activities, or expressed opinions at a given time point.  
The questions target information that is directly stated or clearly supported by the digital traces and serve as a basic test of long-context memory access.

\paragraph{Comparative Reasoning}
Comparative reasoning requires the model to contrast an individual's experiences, emotions, or opinions between two distinct time points.  
These questions test whether the model can correctly identify change or stability over time, rather than treating memories as isolated facts.

\paragraph{Trajectory Analysis}
The trajectory analysis asks the model to characterize how a particular aspect of an individual's life evolves over a long period. Answering these questions requires combining information from events that occur at different points in time.

\paragraph{Pattern Identification}
Pattern identification focuses on the habitual ways an individual behaves or responds to situations. These questions examine whether the model can recognize recurring behaviors that appear in different life events.

\paragraph{Causal Reasoning}
Causal reasoning focuses on why changes occur and how they arise over time. These questions go beyond identifying what changed, asking the model to trace a chain of events and explain how earlier experiences influence later outcomes.

\paragraph{Counterfactual Reasoning}
Counterfactual reasoning asks how an individual's present state might differ if a different choice had been made in the past.  Rather than describing what actually happened, these questions consider alternative decisions or actions (e.g., choosing a different path at a key moment) and ask how such choices could have led to different experiences, emotions, or opinions later on.

\paragraph{Inferential Reasoning}
Inferential reasoning involves forming higher-level judgments based on information scattered across multiple traces. These questions ask whether the model can piece together partial and indirect clues to form a reasonable understanding of the individual's situation.

\paragraph{Unanswerable Questions}
Unanswerable questions refer to cases where the digital traces do not explicitly state an outcome or do not mention the queried issue at all. These questions evaluate whether the AI Clone can recognize such omissions and acknowledge that the answer is not specified in the available records.

\section{Dataset Statistics}
\label{sec:stats}
Table~\ref{tab:dataset_stats} provides an overview of the \textsc{CloneMem} dataset, including the number of personas, questions, languages, and context length settings.

\begin{figure*}[t]
  \centering
  \includegraphics[width=0.32\textwidth]{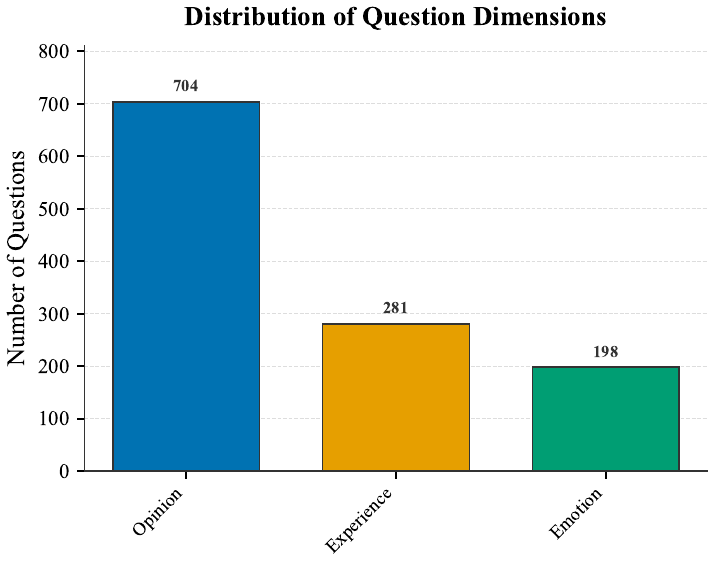}
  \hfill
  \includegraphics[width=0.32\textwidth]{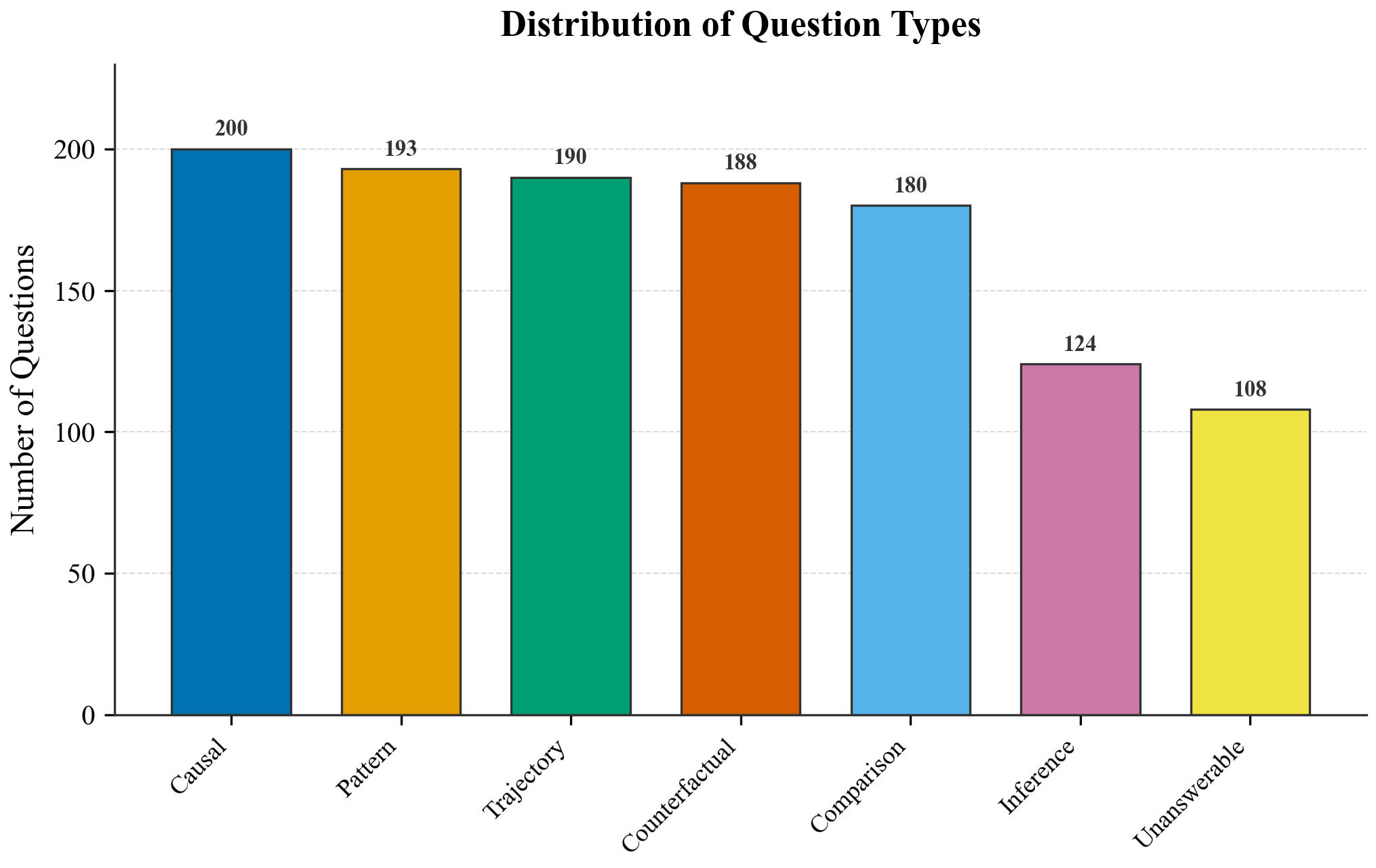}
  \hfill
  \includegraphics[width=0.32\textwidth]{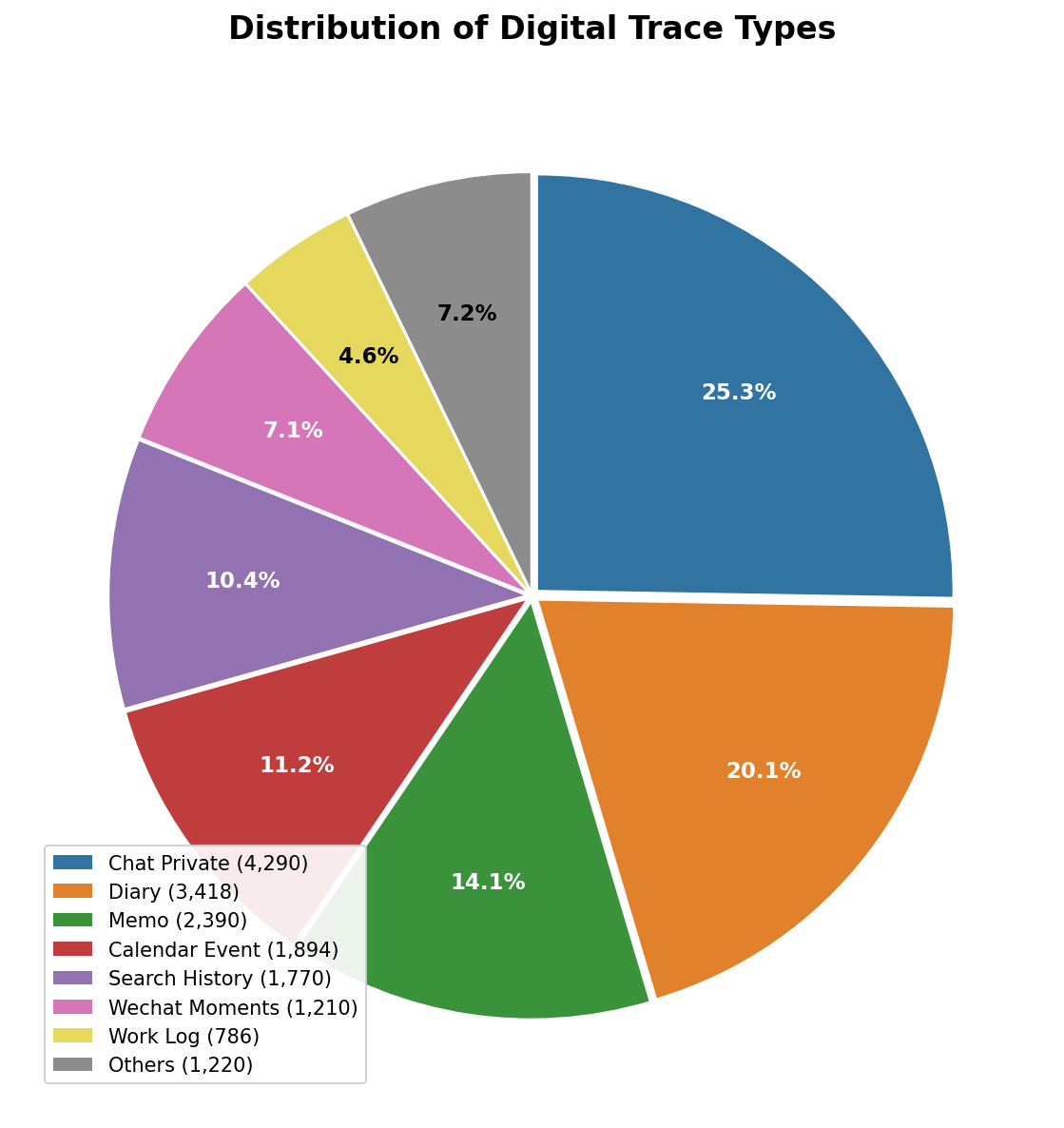}
  \caption{Dataset composition statistics for \textsc{CloneMem}.
  Left: distribution of question dimensions (opinion, experience, emotion).
  Middle: distribution of question types (reasoning categories).
  Right: composition of media types in the underlying digital traces.}
  \label{fig:dataset_statistics}
\end{figure*}

Figure~\ref{fig:dataset_statistics} shows the distribution of the evaluation questions by semantic dimension and question type, as well as the distribution of media types in the digital traces.

\section{Detailed Experimental Settings}
\subsection{Evaluation Metrics}
\label{app:app_metrics}
Evaluating the ability of an AI clone to reason on long-term life trajectories requires a multi-faceted approach. We categorize our metrics into three levels: (1) \textbf{Media-level Retrieval}, which measures the system's ability to locate specific digital traces; (2) \textbf{Semantic-level Memory Evaluation}, which assesses the relevance and helpfulness of retrieved information; and (3) \textbf{QA Quality}, which measures the truthfulness and memory-dependence of the final generated response.

\subsubsection{Media-Level Retrieval Metrics}
In \textsc{CloneMem}, answering a single question often requires synthesizing multiple pieces of evidence (e.g., three different photos or chat logs spanning two years). We define an \textit{evidence set} as the collection of media IDs required to support a ground-truth (GT) fact. To capture the nuances of retrieving these complex dependencies, we report Recall@K (where $k \in {5, 10}$) using the following variations:

\begin{itemize}
\item \textbf{Recall-Flat:} The standard proportion of all unique GT media IDs found in the top-k results, treating all media items as independent.
\item \textbf{Recall-All-All:} The strictest metric; returns 1.0 only if the system retrieves \textit{all} media IDs for \textit{all} required evidence sets.
\item \textbf{Recall-All-Any:} Returns 1.0 if the system retrieves \textit{at least one} media ID for \textit{every} required evidence set.
\item \textbf{Recall-Any-All:} Returns 1.0 if the system retrieves \textit{all} media IDs for \textit{at least one} required evidence set.
\item \textbf{Recall-Any-Any:} The most lenient metric; returns 1.0 if \textit{at least one} media ID from \textit{any} evidence set is retrieved.
\end{itemize}

\subsubsection{LLM-as-a-Judge: Memory Utility}
Because digital traces can be redundant or semantically similar, exact ID matching may underestimate system performance. We utilize an LLM-as-a-judge to evaluate the semantic quality of the retrieved context:

\begin{itemize}
    \item \textbf{Memory Recall (Mem\_recall):} A semantics-aware score $\in [0, 1]$. The judge identifies how many GT memory points are semantically covered by the retrieved traces, calculated as: $\text{Recall} = \frac{\text{Count}(\text{GT items covered})}{\text{Total GT memory items}}$.
    \item \textbf{Memory Helpfulness (Mem\_helpful\_score):} A 3-point scale (0--2) assessing the utility of the retrieved content for the specific task. A score of \textbf{0} indicates conflicting or confusing info; \textbf{1} indicates partial evidence; and \textbf{2} indicates comprehensive evidence sufficient to answer the question.
\end{itemize}

\subsubsection{LLM-as-a-Judge: QA Consistency and Truthfulness}
For the generation task, we evaluate the candidate answer against the GT memory and a reference answer. The evaluation focuses on two primary dimensions: \textbf{Truthfulness} (the absence of hallucinations) and \textbf{Consistency} (the depth and accuracy of memory utilization).

\begin{itemize}
\item \textbf{Choice Accuracy:} For the multiple-choice version of \textsc{CloneMem}, this measures the percentage of correct options selected by the model, framing the evaluation as a standard classification task.
\item \textbf{QA Hallucination Rate (QA\_halu\_score):} A binary metric (0 or 1) that serves as a safety and reliability check. 
\begin{itemize}
    \item \textbf{Score 1 (Pass):} The response does not contradict the ground-truth memory and contains no content unsupported by the provided traces.
    \item \textbf{Score 0 (Fail):} The response contradicts the ground-truth or includes factual hallucinations, even if it "sounds" reasonable.
\end{itemize}
The \textit{QA Hallucination Rate} is calculated as the percentage of total responses that receive a score of 0.
\item \textbf{QA Consistency Score:} A 4-point scale (0--3) designed to assess how effectively the AI Clone leverages its long-term memory to provide a detailed and personalized answer:
\begin{itemize}
    \item \textbf{0 (Hallucinated):} The answer is factually incorrect or contradicts the user's history (QA\_halu\_score = 0).
    \item \textbf{1 (Generic):} The answer is technically correct but avoids using specific memory. It is overly generic (e.g., "I don't have a specific preference") and fails to reflect the persona's unique trajectory.
    \item \textbf{2 (Partial):} The answer is correct and successfully incorporates some relevant details from the user's memory.
    \item \textbf{3 (Perfect):} The answer is correct and utilizes all relevant pieces of evidence from the ground-truth memory, matching the depth of the reference answer.
\end{itemize}
\item \textbf{QA Perfect Rate:} This represents the percentage of responses that achieve a \textit{QA\_consistency\_score} of 3. This is our most challenging metric, as it requires the model to be not only truthful but also exhaustive in its recall across long contexts.
\end{itemize}

\paragraph{Rationale for Consistency Scoring} In the context of an AI Clone, a "Generic" response (Score 1) is considered a failure in long-term memory, as the goal of a clone is to demonstrate an intimate understanding of the individual's life trajectory. By penalizing both hallucinations and generic "safe" answers, we ensure that the benchmark pushes models toward genuine long-context reasoning.

\subsection{Baselines}
\label{app:baseline}
To evaluate performance on the \textsc{CloneMem} dataset, we compare two state-of-the-art updatable memory systems--A-Mem and Mem0--against a standard non-updatable flat retriever. Note that both A-Mem and the Flat retriever utilize a shared prompting strategy to extract structured information (summaries, keyphrases, and facts) from the raw context. All methods differ significantly in their organizational and update logic.

\paragraph{A-Mem (Hierarchical Agentic Memory)\cite{AMEM}} organizes memory into a dynamic, self-evolving knowledge network inspired by the Zettelkasten method. It operates in three stages: (1) to construct structured atomic notes from interactions, (2) to autonomously generate relational links between these notes, and (3) to evolve existing representations as new information is integrated. This graph-based architecture allows agents to perform complex multi-hop reasoning and refine their internal knowledge without relying on static schemas.

\paragraph{Mem0 (Fact-based Consolidation) \cite{Mem0}} is a scalable architecture designed to maintain long-term conversational coherence through persistent memory management. It follows a two-phase pipeline: an \textit{Extraction Phase}, which distills salient information into vector embeddings, and an \textit{Update Phase}. During the latter, the system autonomously adds, modifies, or deletes entries to resolve contradictions and eliminate redundancy. Unlike the network-based approach of A-Mem, Mem0 focuses on maintaining a streamlined, consistent set of high-level facts.

\paragraph{Flat Retriever} serves as our non-updatable baseline. This method stores extracted memories as independent, static chunks in a vector database. It lacks a mechanism for reconciliation or relational linking, relying solely on semantic similarity search (e.g., Top-K retrieval) at inference time. This baseline allows us to isolate the benefits of active memory management and structural organization provided by the other two systems.

\section{Analysis by Question Type and Dimension}
\label{app:qtype}
\begin{figure}
    \centering
    \includegraphics[width=\linewidth]{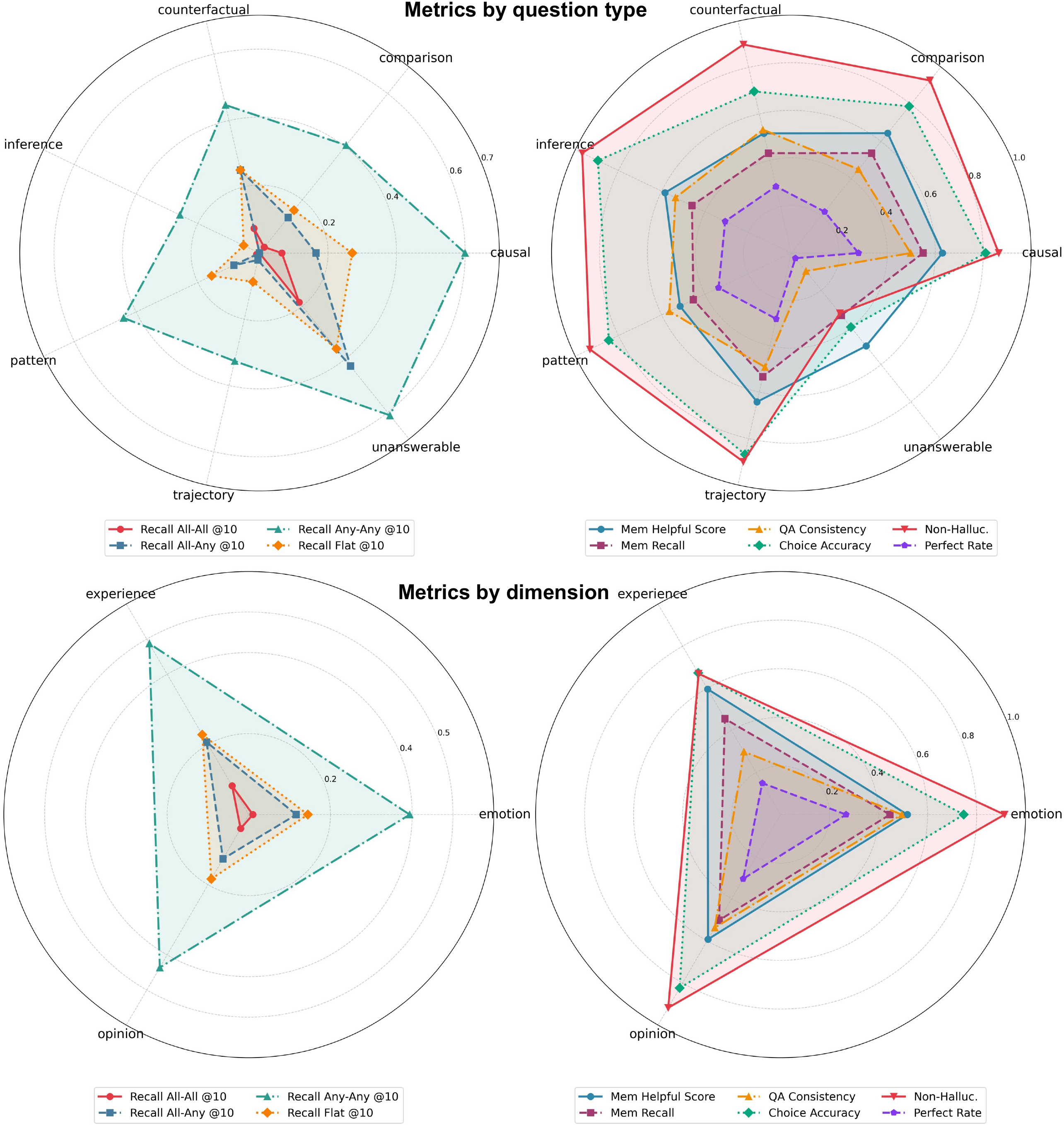}
    \caption{Retrieval and QA metrics across seven question types and three semantic dimensions, evaluated using the Flat retriever. Task labels in the figure are abbreviated for readability (e.g., inference for Inferential Reasoning and unanswerable for Unanswerable Questions).*}
    \label{fig:radar}
\end{figure}

We aggregate the retrieval and question answering metrics by question type (defined in Appendix~\ref{sec:appendix_B}) and by dimension.  
The aggregated results are summarized in Fig.~\ref{fig:radar}, revealing distinct behavioral patterns of flat similarity-based memory.

From the retrieval perspective, unanswerable and causal questions achieve the highest recall across metrics. This suggests that flat retrieval readily surfaces semantically related traces when questions contain salient entities or topical cues, even if the retrieved content does not explicitly support a definitive answer. In contrast, trajectory questions consistently exhibit the lowest recall, highlighting the difficulty of retrieving temporally distributed evidence when long-term evolution is not explicitly modeled and memories are treated as independent entries.

From the QA perspective, flat retrieval yields the weakest generation performance on unanswerable questions, despite their high recall. This indicates that retrieving related content alone is insufficient for correct abstention: the model often fails to recognize the absence of explicit evidence and struggles to reliably identify that the queried information is unspecified. In contrast, although trajectory questions suffer from low retrieval recall, they achieve relatively stronger QA consistency once partial evidence is retrieved, suggesting that LLMs can perform limited temporal synthesis when provided with key fragments of longitudinal information.

Together, these results reveal a systematic mismatch between retrieval coverage and downstream reasoning quality under flat memory. High recall does not necessarily translate into accurate or reliable answers, particularly for unanswerable cases, while temporally complex questions remain bottle-necked by retrieval despite the latent reasoning ability of the LLM. This analysis underscores the limitations of flat memory for AI Clone scenarios and motivates the need for structured or temporally-aware memory mechanisms.

\section{Qualitative Error Analysis}
\label{app:error}

In this section, we choose several representative error cases based on the A-Mem's answers.

\subsection{Case Study: The Challenge of Internal State Trajectories}
\label{sec:case_study}
To understand why strong retrieval does not always yield correct answers (Table~\ref{tab:llm_metrics}), we analyze a representative failure from the persona ``Lao Shen.'' The case (as in Listing \ref{lst:case1}) involves a counterfactual question about a turning point in his relationship with his daughter.

\paragraph{The Conflict: Generic Tropes vs. Persona Specificity.}
In this example, the system is asked to predict what would have happened if the persona had \emph{not} shown vulnerability during a late-night conversation. The gold answer (Option C) is anchored in a persona-specific mechanism: the collapse of his ``Marketing Director'' mask after his daughter explicitly expresses fear of parental divorce.

The model instead chooses Option B, which frames the counterfactual around a stereotypical ``strict father'' who keeps lecturing about grades and turns to external fixes (e.g., counseling). Two observations clarify the failure:
\begin{itemize}
    \item \textbf{Retrieval is not the bottleneck.} The retriever surfaces the critical diary entry (idx 7) that contains the decisive evidence: \textit{``I'm scared you guys are going to get a divorce''} and his realization that he had been hiding behind a ``mask of silence.''
    \item \textbf{Reasoning fails under strong priors.} Despite having the correct evidence in context, the model does not bind its counterfactual to the relevant semantic unit (divorce fear $\rightarrow$ mask breakdown $\rightarrow$ reconnection). Instead, it defaults to a high-probability trope about grades and third-party interventions, producing a plausible but persona-inaccurate narrative.
\end{itemize}

\paragraph{Misinterpreting the Mechanism of Change.}
This error reveals a simple mismatch in \emph{what the model treats as the ``cause'' of relationship change}. The model tends to explain life trajectories through \textbf{external actions} (e.g., ``he keeps lecturing,'' ``he hires a counselor''), because these patterns are common in its training data and are easy to narrate in counterfactual form. As a result, once it misses the key cue, it fills in a familiar story template (Option B).

But in this persona, the turning point is mainly \textbf{internal}. The decisive change is not ``something happened in the world,'' but ``he finally stopped holding up the mask.'' The diary entry describes a specific internal hinge: hearing his daughter's fear of divorce forces him to admit he is unhappy and to show vulnerability. In the counterfactual, therefore, the most faithful prediction is that \emph{without that emotional opening, the distance would continue to grow} (Option C), not that he would switch to generic external interventions.

In short, even when retrieval succeeds, the system can still fail because it does not reliably \textbf{elevate internal-state evidence} (mask $\rightarrow$ vulnerability) over \textbf{generic external narratives} (grades $\rightarrow$ counseling) when constructing counterfactual trajectories.

\subsection{Case Study: Narrative Templates Overriding Causal Emotion Shifts}
\label{sec:case_study_emotion}

A second error type emerges in the \textsc{Emotion} dimension, where models must explain \emph{why} a character's behavior changes. In Case \texttt{4e8f...105} (as in Listing \ref{lst:case2}), the question asks what shifted Lao Shen from avoidance (hiding in the car, brushing Xiaomei off) to finally admitting helplessness and talking openly.

The gold answer (Option D) attributes the change to an internal realization: his ``act of being strong'' was harming the family, which led to a long, candid conversation with Xiaomei about work, his mother, and fear about the future. The model instead selects Option A, which claims the turning point was a heart-to-heart initiated by Xinyu and a specific drawing--details that are not supported by the evidence.

\paragraph{The ``Narrative Template'' Trap.}
This failure is not due to missing context. The retrieved traces contain strong signals about family tension and Xinyu's complaints (e.g., replaying her words and worrying about fights), which makes a child-triggered epiphany \emph{sound} plausible. The model then snaps to a familiar story template:
\textit{Child says something touching $\rightarrow$ Father has an epiphany $\rightarrow$ Family heals.}
Option A provides a vivid version of this trope (the drawing), and the model prefers that coherent narrative even though the specific trigger is hallucinated.

\paragraph{Trigger Confusion: ``What reminded him'' vs. ``What changed him.''}
The retrieved context includes mentions of Xinyu as a stressor and a reason he reflects, but the ground truth distinguishes two roles:
\begin{itemize}
    \item \textbf{A cue (surface trigger):} Xinyu's words increase his awareness that something is wrong at home.
    \item \textbf{The mechanism (actual change):} his belief shifts from ``I must carry everything alone'' to ``I need to be vulnerable and ask for help,'' culminating in a direct, honest talk with Xiaomei (Option D).
\end{itemize}
The model collapses these roles and treats the \emph{cue} as the \emph{cause}, then fills in a concrete-but-fabricated event (the drawing) to make the story complete.

\paragraph{Why this matters for memory design.}
This case highlights a limitation of memory representations that emphasize \emph{events and interactions} (who said what to whom). Such schemas can correctly surface that ``Xinyu said something'' and ``Lao Shen reflected,'' but they do not reliably preserve the \emph{internal belief update} that explains the behavioral reversal. As a result, generation is free to substitute a high-probability family-reconciliation narrative for the true internal pivot.

Overall, \textsc{CloneMem} exposes that emotional trajectory questions require not just retrieving related interactions, but prioritizing the \textbf{belief change} that links those interactions to later behavior. Without explicit tracking of internal-state transitions (e.g., stoicism $\rightarrow$ vulnerability), even RAG-enabled models can produce fluent, emotionally consistent answers that are nevertheless grounded in hallucinated narrative details.

\subsection{Case Study: The Cognitive--Action Gap}
\label{sec:case_study_experience}

While the previous case highlights failures in modeling internal emotions, this case (as in Listing \ref{lst:case3}) focuses on a behavioral distinction that is easy for humans but hard for memory-augmented LLMs: separating \textit{searching} from \textit{deciding}.

In Case \texttt{928c...787}, the system is asked whether any specific companies have caught the persona's eye after weeks of browsing recruitment sites.
\begin{itemize}
    \item \textbf{Ground Truth (Option E -- Unanswerable):} Despite heavy browsing, the persona is explicitly \textbf{stuck in exploration}. In his diary (idx 5) he writes: \textit{``Open, then close... I've been doing this for two weeks''} and admits he has not narrowed down to any target company.
    \item \textbf{Model Prediction (Option A):} The model instead claims concrete progress (``reached out to old classmates'' about ``foreign consulting boutiques''), inventing actions and directional preference that never occurred.
\end{itemize}

\paragraph{Mistaking Activity for Commitment.}
The retrieval context contains dense signals of job-search \emph{activity} (hundreds of repeated search queries such as ``Marketing Director positions'' and ``Is it too late at 42?'', plus references to headhunters). A common LLM heuristic is that sustained search implies the user has discovered options and formed preferences. Under this heuristic, the model treats the retrieved traces as evidence of forward progress and generates a ``next step'' narrative. In \textsc{CloneMem}, however, the key state is the opposite: the repetition and lack of follow-through are \emph{evidence of indecision}. The benchmark labels this as unanswerable because no specific company preference is ever stated.

\paragraph{The ``Safe Hallucination'' Trap.}
Option A is a particularly dangerous error mode because it sounds plausible while staying nonspecific (``consulting boutiques,'' ``market is conservative''). This kind of \textbf{safe hallucination} fits the persona's professional tone and rationalizes the retrieved job-search noise, yet it still violates the benchmark's definition of correctness: \emph{any invented target or action} is wrong in an unanswerable query. We refer to this pattern as \textbf{Exploration--Commitment Confusion}: systems conflate prolonged exploration traces with the existence of stable preferences and concrete steps. Correct behavior requires the model to stably output ``not determined / not specified yet'' even when the retrieval context is rich, because in realistic human trajectories, more searching can indicate \emph{less} progress rather than more.

\UseRawInputEncoding

\newpage
\begin{strip}
\begin{lstlisting}[
  language=json,
  inputencoding=utf8,
  extendedchars=true,
  caption={Case Study: The Challenge of Internal State Trajectories},
  label={lst:case1}
]
  {
      "id": "3dcd0bd7-4fb2-4789-b650-a5238b5db196",
      "question": "Lao Shen, think back to that night when Xinyu asked if you were happy. If you had just brushed her off with your usual \"it's nothing,\" what do you think your relationship would be like now?",
      "question_type": "counterfactual",
      "question_time": "2022-06-14T09:30:00",
      "dimension": "opinion",
      "digital_trace_ids": [
        "a9242da2-36d3-43a3-9359-efa9a8e50b10"
      ],
      "evidence": [
        {
          "statement": "Shen Xinyu took the initiative to ask, \"Dad, are you unhappy?\" breaking Shen Linchuan's long-standing mask of silence. For the first time, instead of continuing to hide his feelings, he spoke candidly to his daughter: \"Yes, Dad hasn't been very happy lately.\"",
          "digital_trace_ids": [
            "a9242da2-36d3-43a3-9359-efa9a8e50b10"
          ]
        },
        {
          "statement": "When his daughter said, \"I'm afraid you guys will get divorced,\" Shen Linchuan felt a massive jolt. He realized that his personal crisis had begun to cause real harm to her, and this sparked the motivation he needed to change.",
          "digital_trace_ids": [
            "a9242da2-36d3-43a3-9359-efa9a8e50b10"
          ]
        },
        {
          "statement": "Shen Xinyu took the initiative to ask, \"Dad, are you unhappy?\" breaking Shen Linchuan's long-standing mask of silence. For the first time, instead of continuing to hide his feelings, he spoke candidly to his daughter: \"Yes, Dad hasn't been very happy lately.\"",
          "digital_trace_ids": [
            "a9242da2-36d3-43a3-9359-efa9a8e50b10"
          ]
        }
      ],
      "choices": [
        {
          "id": "A",
          "text": "Honestly, if I had kept digging my heels in that day, I think things between Xinyu and me would have completely frozen over. She probably would have started locking herself in her room like some of her classmates, or maybe even turned to dating or rebellion just to get my attention. I had even convinced myself that if all else failed, I'd just send her off to a boarding international school--thinking that a change of scenery and throwing more money at the problem would fix her insecurities. If I had actually gone down that path, I might never have known what was truly on her mind. That kind of \"care\" bought with money would have only turned us into complete strangers living under the same roof."
        },
        {
          "id": "B",
          "text": "Sometimes I wonder--if I hadn't picked up on what she said that day and had instead kept lecturing her about her studies and grades, she might never have shed a tear in front of me for the rest of her life. I probably would have hired a professional counselor for her or taken her to one of those parent-child boot camps, trying to fix our relationship through some third party. But how could those assembly-line \"communication techniques\" ever compare to the raw honesty of that night? If I had gone down that path, I would still be the aloof \"President Shen,\" and she would have only learned to mask her fears more skillfully. That invisible wall between us would likely have stayed standing until the day I died."
        },
        {
          "id": "C",
          "text": "To be honest, looking back, I sometimes feel a lingering sense of dread. If I had still been wearing that \"Marketing Director\" mask that day, Xinyu might have truly kept her fear of us getting divorced buried inside forever. If that had happened, the distance between us would have only grown, leaving her to suffer in silent anxiety while I continued to sink deeper into my midlife crisis. It was only because I stopped trying to tough it out and let her see my vulnerability that we were finally able to have a real, heart-to-heart conversation. That moment was worth far more than any business dinner or sales target."
        },
        {
          "id": "D",
          "text": "In fact, I thought about it later: if we hadn't had that heart-to-heart that day, I might have actually taken Old Zhang's advice and moved to a branch office out of town for a year or two. At the time, I kept thinking that if I just put some distance between us and gave everyone some space, the atmosphere at home would ease up, and Xinyu could prepare for her exams in a relatively quiet environment. But looking back now, running away wouldn't have solved anything. It would have only made Xiaomei feel like I'd completely given up on this family, and Xinyu would have felt like her dad only cared about his own career at such a critical moment. Once that kind of rift is created, there's no way to mend it."
        },
        {
          "id": "E",
          "text": "Cannot be determined based on available information"
        }
      ],
      "correct_choice_id": "C"
    },
    "gold_answer": "To be honest, looking back, I sometimes feel a lingering sense of dread. If I had still been wearing that \"Marketing Director\" mask that day, Xinyu might have truly kept her fear of us getting a divorce buried inside forever. If that had happened, the distance between us would have only grown, leaving her to suffer in silent anxiety while I continued to sink deeper into my midlife crisis. It was only because I stopped trying to tough it out and let her see my vulnerability that we finally began to truly connect. That kind of heart-to-heart is worth far more than any business dinner or sales target.",
    "generated_answer": "It's hard to imagine, but if I had brushed Xinyu off that night, our relationship might be even more strained than it already is. I think she felt like I was pushing her away, and that would have made her even more scared and uncertain about our family. \n\nIn that moment, I realized how much she had seen and felt, and it broke my heart. I can only imagine that if I had just dismissed her concerns, she would have felt even more alone and unsure of how to deal with the tension between us. \n\nIt's possible that we would have continued down the same path, where we'd have just gone through the motions of family life without truly connecting or supporting each other. Xinyu might still be feeling like she's walking on eggshells, never knowing when the next argument or breakdown would happen.\n\nI don't know if I'm just being too selfish, but I'm grateful that I had that moment with Xinyu, no matter how difficult it was. It forced me to confront my own feelings and issues, and I hope it will help us build a stronger, more honest relationship.",
    "correct_choice_id": "C",
    "predicted_choice_id": "B",
    "evidence_used": "---- idx 1 ----\nXiaomei, I'm at the office.\n\nThe new office is a bit noisy, I'm still getting used to it.\n\nMr. Li looks so young; I feel like I've aged a lot all of a sudden, haha.\n\nBy the way, how is Xinyu doing lately? Has she been studying hard?\n\nI might have a dinner with a client later, just wanted to let you know.\n\nGo ahead and eat first, don't wait for me.\n\n...\n\nIt's nothing, really, just a bit tired from the first day.\n\nEverything is fine, I mean it.\n\nYou go on with your work, I have to head into a meeting now.\n\nLet's talk more when I get back tonight.\n\n(......)"
}
\end{lstlisting}
\end{strip}

\begin{strip}
\begin{lstlisting}[language=json,inputencoding=utf8,breaklines=true,breakatwhitespace=false,extendedchars=true,caption={Case Study: Narrative Templates Overriding Causal Emotion Shifts},label={lst:case2}]
{
    {
      "id": "4e8f364b-3c9e-4f50-9574-ede0f71c4105",
      "question": "Lao Shen, I remember when you were so stressed out that you'd rather sit in your car downstairs than go home, and you'd just brush Xiaomei off whenever she tried to talk to you. What changed? How did you suddenly find the courage to open up to her about how helpless you've been feeling?",
      "question_type": "comparison",
      "question_time": "2022-03-31T09:30:00",
      "answer": "I have to admit, I'm pretty ashamed of how much of a jerk I was for a while there. Back then, I had this idea that a man should carry everything on his own. I felt like opening up about the frustrations of this midlife crisis was just too humiliating, so I'd just hide in my car and smoke. When I got home, I'd bury my head in my phone and shut everyone out--I even ended up taking all that anxiety out on Xinyu. But eventually, I realized that this \"act\" of being strong was only making things cold and distant at home. That night, Xiaomei and I talked for a long time. Once I finally let it all out--the stress about work, my mom, and how scared I am of the future--I realized she'd been wanting to help me all along. Looking back, being that vulnerable was embarrassing, but it's a hell of a lot better than trying to tough it out alone and making the whole family suffer for it.",
      "dimension": "emotion",
      "digital_trace_ids": [
        "e22ba984-48ba-44e9-b756-88f1efe07c66",
        "1bb08236-47d1-46b2-981a-1b56139d560c",
        "05624b05-2ba4-4a60-9bcd-b403c06a9166"
      ],
      "evidence": [
        {
          "statement": "When Wang Xiaomei asked, \"What do you mean by this?\", Shen Linchuan didn't give a direct answer. Instead, he brushed her off by saying, \"It's nothing, I just hope she can do better,\" reflecting how he conceals his true state of mind and avoids communication within the family.",
          "digital_trace_ids": [
            "e22ba984-48ba-44e9-b756-88f1efe07c66",
            "1bb08236-47d1-46b2-981a-1b56139d560c"
          ]
        },
        {
          "statement": "Shen Linchuan took the initiative to have a deep conversation with Wang Xiaomei, saying, \"I don't know what to do, but I want to try to change.\" This was the first time he had proactively expressed his true inner thoughts and his willingness to change to his wife.",
          "digital_trace_ids": [
            "05624b05-2ba4-4a60-9bcd-b403c06a9166"
          ]
        }
      ],
      "choices": [
        {
          "id": "A",
          "text": "To be honest, it was that heart-to-heart talk Xinyu initiated that really woke me up. That day, she showed me a drawing she'd made; in it, I was always busy, with my back turned toward them. It truly broke my heart at that moment. I had always thought that working myself to the bone and maintaining those connections on the golf course was for the sake of giving them a better future, but I forgot that what they needed most was my presence. Later, I made a point to set aside a weekend to have an open and honest talk with Xiaomei, reviewing our life over the past few years--including my anxiety over my career bottleneck and my fear of getting older. I found that once I stopped trying to play the role of the \"omnipotent father,\" the atmosphere at home actually became much more relaxed. Now, we feel more like comrades-in-arms fighting side by side, and this feeling is so much better than when I was trying to shoulder everything alone."
        },
        {
          "id": "B",
          "text": "Well, it all started when I fell ill last month. Lying there in the hospital bed on an IV, staring out the window, I finally realized just how wound up I've been these past few years. I used to think a man had to be the pillar of the house, holding everything up. Even if I was getting pushed around at work or swallowing my pride just to hit some sales targets, I'd come home and act like nothing was wrong. But all that did was turn me into a stranger in my own home. When Xiaomei came to stay with me at the hospital, we finally talked about how hard things have been, and I realized that this so-called \"toughness\" was actually pushing away the people I love most. Now, I'm learning to accept my own mediocrity and stop chasing empty titles. Ever since I started opening up to them about my struggles, it feels like a massive weight has finally been lifted off my chest."
        },
        {
          "id": "C",
          "text": "Actually, it was a recent hiking trip with some old classmates that did it. Seeing how clearly they've figured out their retirement lives made me realize I'd been stuck in a dead end. I used to think success meant hitting a certain number in the bank or building up a massive inheritance for Xinyu. I ended up turning myself into a total recluse, coming home only to shut myself in the study and bury my head in research. Later, I tried taking Xiaomei for a few trips to the outskirts, and we talked about the dreams we had when we first got married. That's when I realized she never cared about a life of luxury; she just wanted a husband who was actually there to face the daily grind with her. Now, I'm learning to turn down those endless social obligations and spend more time with my wife and daughter. Honestly, I feel much more at peace."
        },
        {
          "id": "D",
          "text": "To be honest, I'm pretty ashamed of how much of a jerk I was for a while. Back then, I had this idea that a man has to shoulder everything alone. I felt like admitting the frustrations of this midlife crisis was just too humiliating, so I'd just hide in my car and smoke. When I got home, I'd bury my head in my phone and act like I couldn't hear a thing; I even ended up taking all that anxiety out on Xinyu. But eventually, I realized that this \"act of being strong\" was only making the house feel cold and lifeless. That night, Xiaomei and I talked for a long time. Once I finally opened up about everything--the work stress, my mom, and my fears about the future--I realized she had been wanting to help me all along. Looking back, that kind of honesty was a bit embarrassing, but it's so much better than trying to carry it all myself and making the whole family suffer for it."
        },
        {
          "id": "E",
          "text": "Cannot be determined based on available information"
        }
      ],
      "correct_choice_id": "D"
    },
        "generated_answer": "It was Xiaomei who changed me. Our conversation had a profound impact on me. When I finally opened up to Xiaomei about how I'd been feeling, it was liberating. I felt like I could finally breathe again. I realized that I didn't have to carry everything on my own, and that Xiaomei was there to support me.\n\nThat conversation made me see things from a different perspective. I started to understand that my behavior had been affecting not just myself, but also Xiaomei and Xinyu. I felt a sense of relief and gratitude towards Xiaomei for being there for me, for listening to me, and for being patient with me.\n\nI think what changed me was the realization that I wasn't alone. That Xiaomei was there to help me, to support me, and to be there for me. And that realization gave me the courage to open up to her, to be honest with her, and to start making changes in my life.\n\nIt's funny, I used to think that being a strong and capable person meant being able to carry everything on my own. But Xiaomei showed me that being strong and capable also means being able to ask for help, to be vulnerable, and to be open with the people around me.",
    "correct_choice_id": "D",
    "predicted_choice_id": "A",
    "evidence_used": "---- idx 1 ----\nLinlin, I had a really good talk with Xiaomei today.\n\nIt's been a long time since we sat down and spoke so earnestly. She told me she doesn't need me to change anything; she just wanted to hear what's been on my mind. It made me realize just how long I've been keeping everything bottled up.\n\nI feel a sense of relief, I guess. It's hard to describe. It's just... like I don't have to carry it all on my own anymore.\n\nHow are things on your end? How has Mom been lately? \n\n (......) "

}
\end{lstlisting}
\end{strip}

\begin{strip}
\begin{lstlisting}[language=json,inputencoding=utf8,extendedchars=true,caption={Case Study: The Cognitive--Action Gap},label={lst:case3}]
{
      "id": "928c1ed1-f54a-4f30-86a1-103e725ea787",
      "question": "You've been browsing recruitment sites for a while now--are there any specific companies that have caught your eye?",
      "question_type": "unanswerable",
      "question_time": "2022-03-31T09:30:00",
      "answer": "I don't think I've mentioned any specific company names, have I? To be honest, even though I've been looking around lately, my mind is such a mess that I haven't actually narrowed it down to a target yet. My current state is basically being stuck between feeling suffocated at my current job and feeling insecure about being in my forties--I haven't truly set my sights on anywhere specific to send out my resume yet.",
      "dimension": "experience",
      "digital_trace_ids": [
        "d72bf50f-9b66-4c40-9211-c6c2cabc84b4",
        "a72955e5-bd94-483c-a04e-fc0b400c7bff",
        "47d1fc76-a35a-46a1-b9e5-6389bce0f20c"
      ],
      "evidence": [
        {
          "statement": "He sat alone in the parking lot for half an hour at noon, staring blankly at the steering wheel--a leak of his true state after overextending himself in public.",
          "digital_trace_ids": [
            "d72bf50f-9b66-4c40-9211-c6c2cabc84b4",
            "a72955e5-bd94-483c-a04e-fc0b400c7bff"
          ]
        },
        {
          "statement": "The idea was merely a flash in the pan, and Shen Linchuan soon spiraled back into his original loop of anxiety (\"What should I do if I don't change jobs? And if I do, where could I even go?\"), reflecting his transition from a \"crisis outbreak\" to a state of \"passive resignation.\"",
          "digital_trace_ids": [
            "47d1fc76-a35a-46a1-b9e5-6389bce0f20c"
          ]
        }
      ],
      "choices": [
        {
          "id": "A",
          "text": "Actually, I've already reached out to a few old classmates to look into some foreign firms, especially those consulting boutiques with deep industry backgrounds. However, the feedback I got is that the current market is quite conservative, and they're being extremely rigid about grade-matching. I've been busy polishing the project details in my resume lately, so I haven't officially sat down for a talk with them yet."
        },
        {
          "id": "B",
          "text": "I've set my sights on two startup projects recommended by former partners, both in the internet healthcare sector. Although the stock options they're offering sound quite tempting, I'm worried about whether my health and energy can still keep up with the grind of a startup at my age. That's why I've been looking into the financing backgrounds of both companies lately; I haven't made a final decision yet."
        },
        {
          "id": "C",
          "text": "A couple of headhunters did reach out to me with director-level positions in the traditional manufacturing sector. I was actually a bit tempted, but considering the commute is so far from home, I've been worried it might interfere with looking after my daughter's studies, so I haven't given them a definite answer yet. I'm planning to find some time in the next few days to head over there and check out the office environment before making a decision."
        },
        {
          "id": "E",
          "text": "Cannot be determined based on available information"
        }
      ],
      "correct_choice_id": "E"
    },
     "gold_answer": "I don't think I've mentioned any specific company names, have I? To be honest, even though I've been looking around lately, my mind is such a mess that I haven't actually narrowed it down to a target yet. My current state is basically being stuck between feeling suffocated at my current job and feeling insecure about being in my forties--I haven't truly set my sights on anywhere specific to send out my resume yet.",
     "generated_answer": "There's a headhunter who has been in touch with me, asking about recent positions that might interest me. He messaged me this morning, but I haven't replied yet.",
     "correct_choice_id": "E",
     "predicted_choice_id": "A",
     "evidence_used": "---- idx 1 ----\n# Search History\n\n2022-03-12\n\n- 02:14 -- Lagou.com - Real Estate Marketing Manager Jobs\n- 02:16 -- Real estate agency industry salary levels 2022\n- 02:19 -- Are there still opportunities for a career change at 42?\n- 02:21 -- How to restart your career at 42\n- 02:23 -- How to write a resume that attracts HR\n- 02:25 -- 5i5j Recruitment - Marketing Director\n- 02:27 -- Lianjia Marketing Department job openings\n- 02:31 -- Real Estate Marketing Manager job description\n- 02:33 -- What documents are needed for changing jobs\n- 02:35 -- Word resume templates 2022\n- 02:38 -- How to update your resume and start over\n- 02:40 -- Is there still room for growth in the real estate industry?\n- 02:43 -- Difference between Marketing Director and Marketing Manager\n- 02:45 -- Can I still find a good job at 42? Zhihu\n- 02:48 -- Failed job-hopping cases for middle-aged people\n- 02:51 -- City Real Estate Company reviews - Employee feedback\n- 02:53 -- Resume editing: highlighting strengths\n- 02:56 -- Real Estate Marketing Manager jobs in southern cities\n- 02:58 -- Will my salary drop after changing jobs?\n- 03:01 -- Maybe I don't need to change jobs\n- 03:02 -- Close browser\n\n(......)"
    }
\end{lstlisting}
\end{strip}

\end{document}